\documentclass[journal]{IEEEtran}
\usepackage{graphicx}
\usepackage{amsmath}
\usepackage{amssymb}
\usepackage{pifont}
\usepackage{xcolor}
\usepackage{bm}
\usepackage{cite}
\usepackage{multicol}
\usepackage{multirow}
\usepackage{array}
\usepackage{url}
\usepackage{booktabs}
\usepackage{soul}
\usepackage[inline]{enumitem}
\usepackage{arydshln}
\usepackage{xspace}

\usepackage{siunitx}
\usepackage{xstring}

\newcommand{\PotsdamNoClutterTinyOA}{93.53}
\newcommand{\PotsdamNoClutterTinymFscore}{93.7}
\newcommand{\PotsdamNoClutterTiny}{88.45}
\newcommand{\PotsdamNoClutterSmall}{88.6}

\newcommand{\PotsdamWithClutterTiny}{79.5}
\newcommand{\PotsdamWithClutterSmall}{79.3}
\newcommand{\PotsdamWithClutterBase}{79.7}%
\newcommand{\iSAIDTiny}{67.5}
\newcommand{\iSAIDSmall}{68.4}
\newcommand{\iSAIDBase}{69.3}
\newcommand{\LovedaTestTiny}{52.0}
\newcommand{\LovedaTestSmall}{52.4}
\newcommand{\LovedaTestBase}{54.1}

\newcolumntype{C}{>{\centering\arraybackslash}p{37pt}}
\newcolumntype{A}{>{\centering\arraybackslash}p{43pt}}

\definecolor{Azure}{RGB}{78, 103, 176} %
\newcommand{\score}[3][black]{%
  \begingroup
  \StrLen{#2}[\length]%
  \ifnum\length>#3
    \textcolor{#1}{\num[round-mode=places, round-precision=(#3-2)]{#2}}%
  \else
    \textcolor{green}{#2}%
  \fi 
  \endgroup
}

\newcommand{\secondplace}[1]{\underline{\textit{#1}}}

\usepackage{subcaption}
\usepackage{textcomp}
\usepackage{algpseudocode}

\begin{document}

\newcommand{\model}{AerialFormer\xspace}

\title{\model: Multi-resolution Transformer for Aerial Image Segmentation}

\author{ Kashu~Yamazaki$^*$,~\IEEEmembership{Member,~IEEE,} Taisei~Hanyu$^*$, Minh Tran, Adrian de Luis,
Roy McCann, Haitao Liao, Chase Rainwater,  Meredith Adkins, 
Jackson Cothren, and
Ngan~Le,~\IEEEmembership{Member,~IEEE}%
\thanks{K. Yamazaki, T. Hanyu, M. Tran, A. Luis, and N. Le are with Artificial Intelligence and Computer Vision (AICV) Lab, University of Arkansas, 1 University of Arkansas
Fayetteville, AR 72701 USA e-mail: \{kyamazak, thanyu, minht, ad084, thile\}@uark.edu}%
\thanks{J. Cothren is with Department of Geoscience, Center for Advanced Spatial Technologies, University of Arkansas, 227 N Harmon Ave, Fayetteville, AR 72701, jcothre@uark.edu}
\thanks{R. McCann is with Department of Electrical Engineering, University of Arkansas, 1 University of Arkansas
Fayetteville, AR 72701, rmccann@uark.edu}
\thanks{H. Liao and C. Rainwater are with Department of Industrial Engineering, University of Arkansas, 1 University of Arkansas
Fayetteville, AR 72701, {liao, cer}@uark.edu}
\thanks{M. Adkins is with Institute for Integrative and Innovative Research, University of Arkansas, 1 University of Arkansas Fayetteville, AR 72701, mmckee@uark.edu}

}

\markboth{IEEE Transactions on Geoscience and Remote Sensing, 2023}%
{Yamazaki and Hanyu\MakeLowercase{\textit{et al.}}: \model: Multi-resolution Transformer for Aerial Image Segmentation}

\maketitle

\begin{abstract}
When performing Aerial Image Segmentation, practitioners  often encounter various challenges, such as a strong imbalance in the foreground-background distribution, complex background, intra-class heterogeneity, inter-class homogeneity, and the presence of tiny objects. To overcome these challenges by leveraging the power of Transformers, we propose \model to unify Transformers at the contracting path with lightweight Multi-Dilated Convolutional Neural Networks (MD-CNNs) at the expanding path. The proposed \model is designed with a hierarchical structure, in which the Transformer encoder generates multi-scale features and the MD-CNNs decoder aggregates the information from the multi-scale inputs. As a result, the information in both local and global contexts is taken into consideration, so that powerful representations and high-resolution segmentation can be achieved. We have tested the proposed \model on three benchmark datasets, including iSAID, LoveDA, and Potsdam. Comprehensive experiments and extensive ablation studies show that the proposed \model remarkably outperforms those state-of-the-art methods. Our source code will be publicly available upon acceptance. 
\end{abstract}


\IEEEpeerreviewmaketitle

\begin{figure*}[t]
\centering
\includegraphics[width=\linewidth]{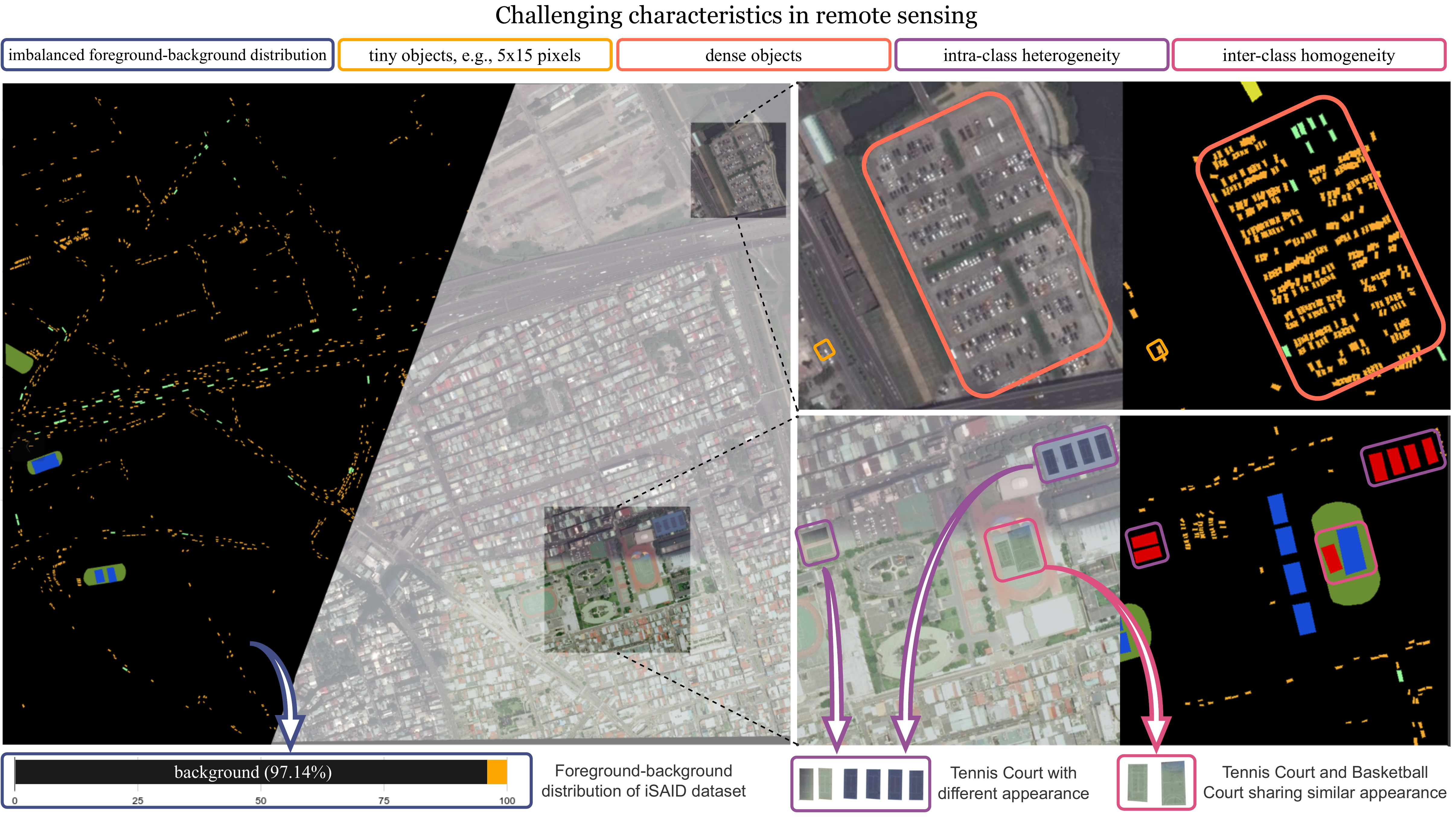}
\caption{Examples of challenging characteristics in remote sensing image segmentation. (left) the distribution of foreground and background is highly imbalanced. (right-top) objects in some classes are dense and small such that they are hardly identifiable. (right-bottom) within a class, there is a large diversity in appearance: intra-class heterogeneity (purple); some different classes share the similar appearance: inter-class homogeneity (pink).
The image is from iSAID dataset. Best viewed with color and zoom-in.}
\vspace{-0.5cm}

\label{fig: critical_issue}
\end{figure*}

\section{Introduction}
\label{sec:intro}
The use of aerial images provides a view of the Earth from above, which consists of various geospatial objects such as cars, buildings, airplanes, ships, etc., and allows us to regularly monitor some large areas of the planet. The recent advances in sensor technology have promoted the potential use of those remote sensing (RS) images in broader applications thanks to the ability to capture high spatial resolution (HSR) images with abundant spatial details and rich potential semantic content. Aerial image segmentation (AIS) is a particular semantic segmentation task that aims to assign a semantic category to each image pixel. Thus, AIS plays an important role in the understanding and analysis of remote sensing data, offering both semantic and localization cues for the targets of interest.
Understanding and analyzing these objects from 
a top-down perspective offered by remote sensing (RS) imagery is crucial for urban monitoring and planning. 
This understanding finds utility in numerous practical urban-related applications,
such  as disaster monitoring \cite{schumann2018assisting}, agricultural planning \cite{weiss2020remote}, street view extraction \cite{griffiths2019improving, shamsolmoali2020road}, land change \cite{samie2020examining, marcos2018land, xia2023openearthmap}, land cover \cite{wang2021loveda}, climate change \cite{o2013use}, deforested regions \cite{andrade2020evaluation}, etc. However, due to the large footprint of aerial images and limited sensor bandwidth, several challenging characteristics are needed to be investigated. Some critical issues are intra-class heterogeneity (i.e., objects of the same category may be shown in various shapes, textures, colors, scales, and structures), inter-class homogeneity (i.e., objects of the different classes may share the same visual properties) \cite{wang2020cse}, the large diversity of resolution and orientation \cite{Waqas2019}, dense and tiny objects \cite{shafique2022deep}, background and foreground imbalance \cite{zheng2020foreground}, high background complexity \cite{zheng2020foreground}. As shown in Figure\ref{fig: critical_issue}, the ratio between foreground and background is 2.86\%/97.14\%; the inter-class homogeneity is presented by Tennis Court and Basketball Court, which share similar appearance; intra-class heterogeneity is presented by Tennis Court which are shown in various appearances.

With the success of deep learning (DL) techniques in extracting rich contextual feature e.g., VGG \cite{simonyan2014very}, ResNet \cite{he2016deep}, InceptionNet \cite{szegedy2016rethinking, szegedy2017inception}, MobileNet \cite{howard2017mobilenets}, etc., various semantic segmentation approaches based on those backbones have been proposed such as Unet \cite{ronneberger2015u}, PSPNet \cite{zhao2017pyramid}, DeepLabV3+ \cite{chen2018encoder}, Segmenter \cite{strudel2021segmenter}, or UperNet \cite{xiao2018unified}. Most of the existing image segmentation methods are originally proposed for other use cases such as self-driving vehicle \cite{minaee2021image} and medical imaging \cite{le2021narrow}. Thus, they do not perform optimally on AIS, resulting in limited accuracy on tiny objects and weak boundary objects. To alleviate those limitations, it is essential to obtain strong semantic representations at both the local level (e.g., boundary) and the global context level (e.g., the relationship between objects/classes).

Recently, the great success of Transformer \cite{vaswani2017attention} in natural language processing (NLP) has inspired numerous tasks in computer vision including semantic segmentation. Following the Transformer design in NLP, \cite{dosovitskiy2020image} split an image into multiple linearly embedded patches and feed them into a standard Transformer, and proposes Vision Transformer (ViT) for image classification. Later, \cite{zheng2021rethinking} adopts ViT as a backbone and proposes SETR to demonstrate the feasibility of using Transformers in semantic segmentation. Although the Transformer-based encoder has various benefits, its computational complexity is considerably greater than that of the CNN-based encoder because of its self-attention mechanism with a squared complexity. As a result, it is challenging to process high-resolution images using Transformer-based models. To reduce the computational complexity, some Transformer models such as Swin Transformer \cite{liu2021swin} propose shifted windows to bring greater efficiency by limiting self-attention computation to non-overlapping local windows while also allowing for cross-window connection. 
Despite the great potential in various computer vision tasks owing to their strong capability to model long-range dependency using the self-attention mechanism, vision transformers are limited in modeling local visual structures and scale-invariant representations in the context of dense prediction tasks. Unlike vision transformers, Convolution Neural Networks (CNNs) are based on convolution to compute local correlation among neighbor pixels. Consequently, CNNs are good at extracting local features, and scale invariance and still serve as prevalent backbones in vision tasks. Generally, CNNs and vision transformers focus on different aspects. On one hand, CNNs adopt convolutions allowing CNNs to preferably extract local contextual information and translation invariance. However, this property leads to locality and strong inductive biases. On the other hand, vision transformers adopt self-attention mechanisms for perfectly extracting global and long-range dependencies, but do not capture locality and translation invariance very well. According to the above-mentioned analyses, we believe CNNs and vision transformers are naturally complementary to each other. Thus, combining these two kinds of CNNs and vision transformers can overcome the weaknesses of the two models and strengthen their advantages simultaneously.

In an effort to mitigate the multiple aforementioned challenging characteristics involved in aerial image segmentation, as per our prior analysis, we draw inspiration from the strengths and success of CNNs for exploring the advantages of introducing local visual structures, as well as from the scale-invariant representation in vision transformers.
In this paper, we particularly propose \model, a deep learning network with Swin Transformer encoder and CNNs decoder to efficiently localize objects in aerial images from satellite. Furthermore, we present a new approach that 
utilizes a convolutional stem network to generate fine feature maps for tiny objects in the encoder. We also introduce a Multi-Dilated Convolution (MDC) block at the decoder to effectively extract features while avoiding excessive computational complexity due to its fully convolutional design.

Our contribution is summarized as follows:
\begin{itemize}
    \item Provide a comprehensive literature review on aerial image segmentation.
    \item Analyze the current challenging characteristics of aerial image segmentation.
    \item Propose an effective computation model to leverage the merits of both vision transformers to capture long-range\newpage dependency and CNNs to extract local representation and scale-invariance. 
    \item Propose a CNN stem network to alleviate the potential drawback of using Transformer backbone for dense prediction tasks.
    \item Conduct an extensive experiment on the widely recognized three datasets: iSAID, LoveDA, and Potsdam.
\end{itemize}

\section{Related Works}

Generally, image segmentation is categorized into three tasks: instance segmentation, semantic segmentation, and panoptic segmentation. Each of these tasks is distinguished based on their respective semantic considerations.
In this work, we focus on the second task of semantic segmentation, a form of dense prediction task where each pixel from an image is associated with a class label. Different from instance segmentation, it does not distinguish each individual instance of the same object class. The goal of semantic segmentation is to divide an image into several visually meaningful or interesting areas for visual understanding according to semantic information. Semantic segmentation plays an important role in a broad range of applications, e.g., scene understanding, medical image analysis, autonomous driving, video surveillance, robot perception, satellite image segmentation, agriculture analysis, etc. 
We start this section by reviewing on DL-based semantic image segmentation and the advancements made in Computer Vision with Transformers. Then, we turn our focus to a review of aerial image segmentation using deep neural networks.

\subsection{DL-based Image Segmentation}

Convolutional Neural Networks (CNNs) are widely regarded as the de-facto standard for various tasks within the field of computer vision.
Long et al. \cite{long2015fully} show that Fully Convolutional neural (FCNs) can be used to segment images without fully connected layers and it has become one of the principal networks for semantic segmentation.
With the advancements brought by the FCNs in semantic segmentation, many improvements have been made by designing the network deeper, wider, or more effective. This includes enlarging the receptive field \cite{chen2017deeplab, dai2017deformable, chen2018encoder, yang2018denseaspp, hoang2022dam, le2021multi}, strengthening context cues \cite{le2018deep, he2019adaptive, le2021multi}, \cite{hsiao2021specialize,
hu2020class, jin2021mining, jin2021isnet, yu2020context, yuan2019segmentation, zhang2018context} 
leveraging boundary information \cite{bertasius2016semantic, le2021offset, ding2019boundary, le2021narrow, li2020improving, zhen2020joint}, and incorporating neural attention \cite{harley2017segmentation, he2019dynamic, zhao2018psanet, hu2018squeeze, huang2019ccnet, li2018pyramid, sun2020mining, wang2021hierarchical, wang2018non}.
Recently, a new paradigm of neural network architecture that does not employ any convolutions and mainly relies on a self-attention mechanism, called Transformers, has become rapidly adopted to CV tasks \cite{carion2020end, liu2022dab, li2022dn} and achieved promising performance.  
The core idea behind transformer architecture \cite{vaswani2017attention} is the self-attention mechanism to capture long-range relationships. 
In addition, Transformers can be easily parallelized, facilitating training on larger datasets. 
Vision Transformer (ViT) \cite{dosovitskiy2020image} is considered one of the first works that applied the standard Transformer to vision tasks. Unlike the CNNs structure, the ViT processes the 2D image as a 1D sequence of image patches. Thanks to the powerful sequence-to-sequence modeling ability of the Transformer, ViT demonstrates superior characterization of extracting global context, especially in the lower-level features compared to the CNN counterparts. 
Recent advancements in Transformers over the past few years have demonstrated their effectiveness as backbone networks for visual tasks, surpassing the performance of numerous CNN-based models trained on large datasets. 
Transformer-based image segmentation approaches \cite{cheng2021per, strudel2021segmenter, zheng2021rethinking, xie2021segformer, tran2022aisformer, cheng2022masked} 
inherit the flexibility of Transformers in modeling long-range dependencies, yielding remarkable results. 
Transformers have been applied with notable success across a variety of computer vision tasks. These include image recognition \cite{dosovitskiy2020image, touvron2021training} object detection \cite{carion2020end, zhu2020deformable, sun2021sparse}, image segmentation \cite{ye2019cross, zheng2021rethinking, tran2022aisformer}, action localization \cite{vo2021aei, vo2022aoe}, and video captioning \cite{yamazaki2022vlcap, yamazaki2022vltint}, thereby showcasing their capability to augment global information.

\subsection{Aerial Image Segmentation}
Computer vision techniques have long been employed for the analysis of satellite images. 
Historically, satellite images had a lower resolution and the goal of segmentation was primarily to identify boundaries like straight lines and curves in aerial pictures. However, modern satellite imagery possesses significantly higher resolution, and consequently, the demands of segmentation tasks have substantially increased, which include the segmentation of tiny objects, objects with substantial scale variation, and entities exhibiting visual ambiguities.
To this end, FCNs and their variants have become the mainstream solution for aerial image segmentation and led to state-of-the-art performance across numerous datasets \cite{chen2018encoder, sun2019deep, li2021pointflow, xue2022aanet, factseg2022, hou2022bsnet}.
To capture contextual interrelations among pixels in remote sensing images, techniques from natural language processing have also been incorporated into aerial image segmentation \cite{you2019pixel}.
By imitating the channel attention mechanism \cite{hu2018squeeze}, S-RA-FCN \cite{mou2020relation} designs a spatial relation module to capture global spatial relations, and \cite{niu2021hybrid} introduces HMANet with spatial interaction while balancing between the size of the receptive field and the computation cost. In HMANet, a region shuffle attention module is proposed to improve the efficiency of the self-attention mechanism by reducing redundant features and forming region-wise representations. 
In recent years, the advancements in transformer-based networks, which leverage self-attention mechanisms to achieve receptive fields as large as the entire image, have sparked increased interest in their applications. 
Consequently, there has been a surge in research studies 
\cite{wang2022empirical, wang2022novel, xu2023rssformer, xie2021segformer, wang2022unetformer, chen2021transunet, sun2022ringmo} 
that have integrated Transformers into remote sensing applications. For instance, 
RSSFormer \cite{xu2023rssformer} proposed the Adaptive Transformer Fusion Module to mitigate background noise and enhance object saliency during the fusion of multi-scale features. Some other works \cite{wang2022novel, sun2022ringmo} adopt Transformers as their backbone.  

\begin{figure*}[!t]
\centering
\includegraphics[width=\linewidth]{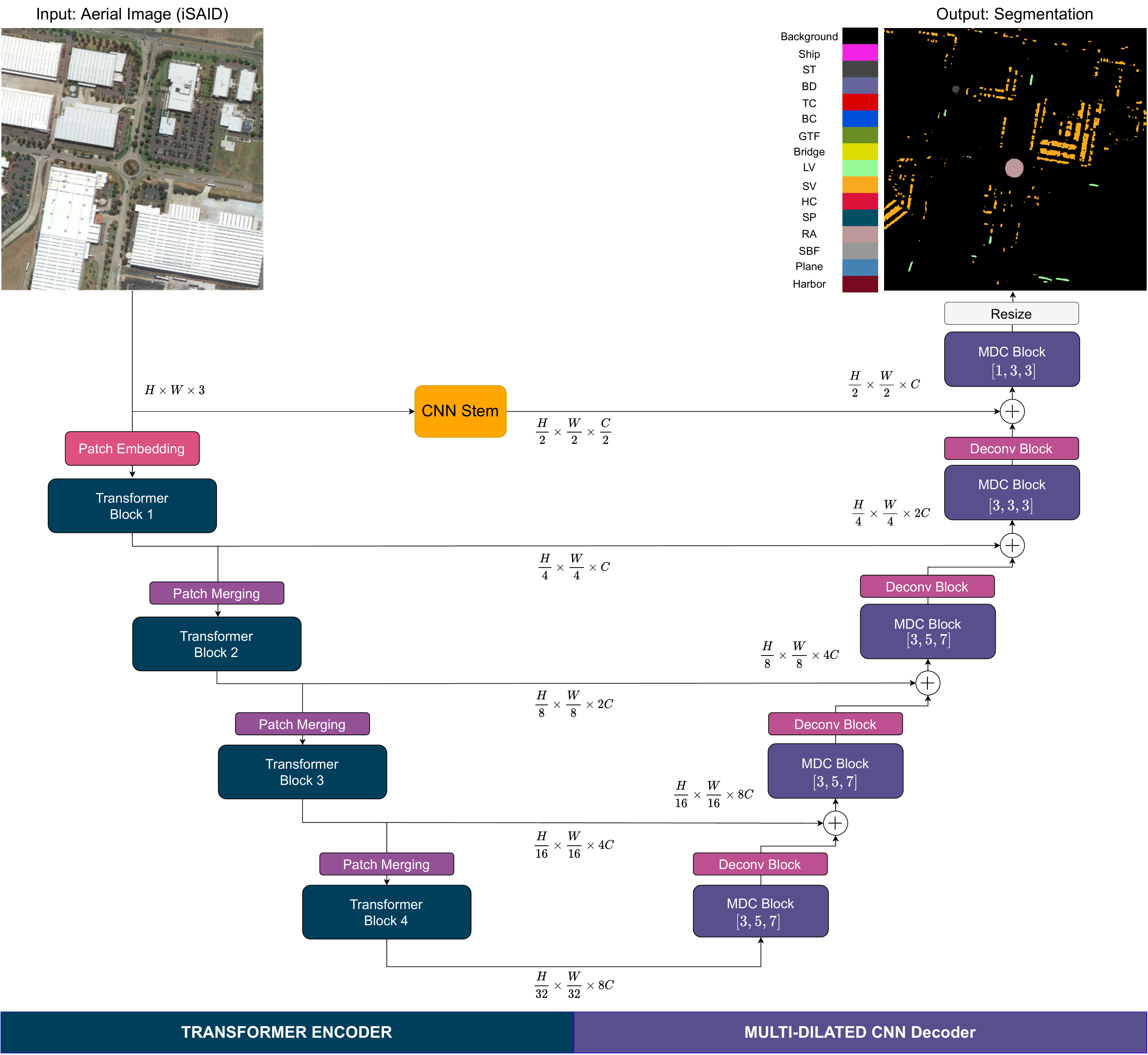}
\caption{Overall network architecture of our proposed \model which consists of three components i.e., Transformer Encoder, CNNs Stem, and Multi-Dilated CNNs Decoder.}
\label{main_fig}
\end{figure*}

In this paper, we introduce \model, an innovative fusion of a Transformer encoder and a multi-dilated CNNs decoder. 
While Transformer-based approaches excel at modeling long-range dependencies, they face challenges in capturing local details and struggles in handling tiny objects. Thus, our \model incorporates multi-dilated convolution to capture long-range dependence without increasing the memory footprint at the decoder. Our novel \model approach combines the strengths of a Transformer encoder and a multi-dilated CNNs decoder, aided by skip connections, to capture both local context and long-range dependencies effectively in aerial image segmentation.

\section{Methods}
\subsection{Network Overview}
An overview of our \model architecture is presented in Figure \ref{main_fig}. The architecture design is fundamentally rooted in the renowned Unet structure for semantic segmentation \cite{ronneberger2015u}, characterized by its encoder-decoder network with use of skip-connections between the matched blocks with identical spatial resolution on both encoder and decoder sides. The composition of our model is threefold: a\textit{ Transformer Encoder}, a \textit{CNNs Stem}, and a \textit{Multi-Dilated CNNs Decoder}. The Transformer Encoder is designed as a sequence of $s$ stages of Transformer Encoder blocks ($s$ is set as 4 in our architecture) aiming to extract long-range context representation. The CNNs Stem aims to preserve low-level information at high resolution. The latter, Multi-Dilated CNNs (MDC) Decoder consists of  $s+1$ MDC blocks with skip connections to obtain information from multiple scales and wide contexts. We will detail these components in the following subsections. 

Given a high-resolution aerial image, we first overlap partition it into a set of sub-images sized $H \times W \times 3$, where 3 corresponds to three color channels. Each sub-image is then fed to the AerialFormer and the output is the segmentation of $H \times W$.

\subsection{Transformer Encoder}
\label{TransEnc}

The Transformer Encoder starts by processing an input image size of $H \times W \times 3$, which is tokenized by the \textit{Patch Embedding layer}, which results in a feature map $\frac{H}{p} \times \frac{W}{p} \times C$. The feature map is then passed through a sequence of $s = 4$ \textit{Transformer Encoder Blocks} and produces multi-level outputs of different sizes at each block: $\frac{H}{4} \times \frac{W}{4} \times C$, $\frac{H}{8} \times \frac{W}{8} \times 2C$, $\frac{H}{16} \times \frac{W}{16} \times 4C$, and  $\frac{H}{32} \times \frac{W}{32} \times 8C$. Each Transformer Encoder Block is followed by a \textit{Patch Merging layer}, which reduces the spatial dimension by half before being passed to the next deeper Transformer Encoder Block.

\subsubsection{Patch Embedding}
The Transformer Encoder starts by taking an image $H \times W \times 3$ as input and dividing it into patches of size $p \times p$ in a non-overlapping manner. Each patch is embedded into a vector in dimensional space of $\mathbb{R}^C$ by a linear projection, which can be simplified as a single convolution operation with the kernel size of $p \times p$ and the stride of $p \times p$. The Patch Embedding produces feature maps of $\frac{H}{p} \times \frac{W}{p} \times C$.
The patch size determines the spatial resolution of the input sequence of the transformer, and therefore smaller patch size is favored for the dense prediction tasks including semantic segmentation. While ViT \cite{dosovitskiy2020image} is a commonly used vision transformer in computer vision, which  processes $16 \times 16$ patch and is able to capture a wider range context, it may not be suitable for capturing detailed information. One of the most challenging aspects of aerial image segmentation is dealing with tiny objects. On the other hand, Swin Transformer \cite{liu2021swin}, one of the transformer variants utilizes a smaller patch of $4\times 4$. Thus, we adopt Swin Transformer \cite{liu2021swin} to implement a Patch Embedding layer to better capture the detailed information of tiny objects in aerial image segmentation.

\subsubsection{Transformer Encoder Block}

\begin{figure}
    \centering
    \includegraphics[width=\linewidth]{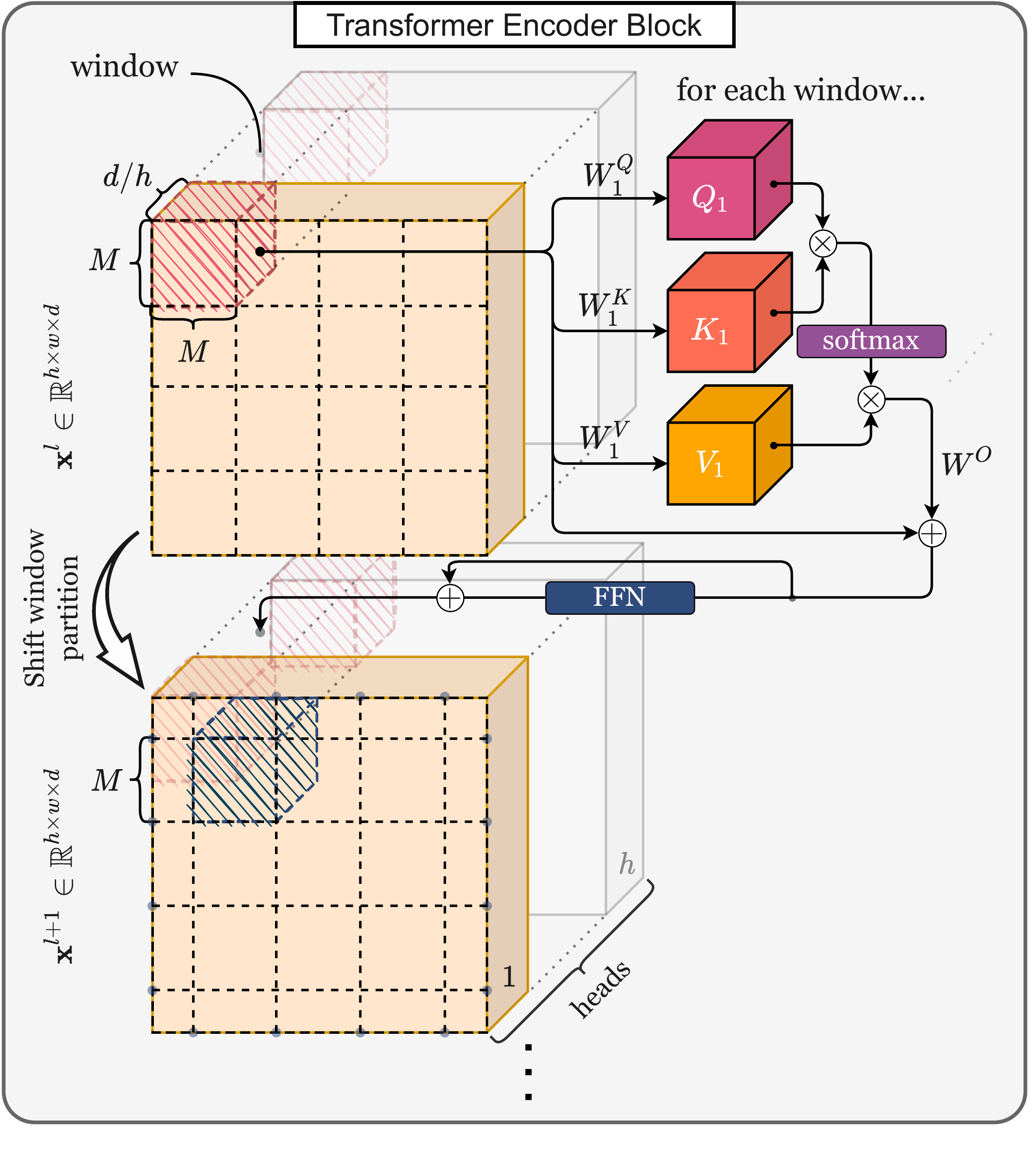}
    \caption{An illustration of the Transformer Encoder Block. }
    \label{fig:enc_block}
\end{figure}

In general, let $\textbf{x} \in R^{h \times w \times d}$ denote the input of a Transformer Encoder Block. Transformer Encoder Block processes the input data with a series of self-attention and feed-forward network with residual connection. To compensate the increase in computation because of the smaller patch size, 
Swin Transformer \cite{liu2021swin} utilizes a local self-attention instead of global self-attention. The global self-attention, used in standard Transformers, has a computational cost of $\mathbb{O}(N^2\cdot d)$ where $N$ is the number of tokens (i.e., $N = h \times w$) and $d$ is the representation dimension, which can be prohibitively expensive for large images and small patch size. 
Swin Transformer introduced window-based self-attention (WSA) that divides the image into non-overlapping windows and performs self-attention within each window. With WSA, the computational cost is linear to the number of tokens, i.e., $\mathbb{O}(M^2\cdot N \cdot d)$ where $M^2$ is the number of patches within a window and $M^2 \ll N$. 
In order to apply the WSA, an input  $\mathbf{x} \in \mathbb{R}^{h\times w \times d}$ is partitioned into a group of local patches $\mathbf{x}' \in \mathbb{R}^{\frac{h\times w}{M^2} \times M^2 \times d}$ and the first dimension $\frac{h\times w}{M^2}$ is treated as a batch dimension, i.e., the network parameters are shared along the first dimension. 
Considering the multi-head attention operation with $h$ heads, the feature dimension $d$ is split into $h$ identical blocks, i.e., $\mathbb{R}^{\frac{h\times w}{M^2} \times M^2 \times \frac{d}{h} \times h}$. Then, we can formulate the WSA as:

\begin{equation}
    \operatorname{WSA}(\mathbf{x}') = [\operatorname{head}_{1}; \dots; \operatorname{head}_{h}] W^O
\end{equation}

where $[;]$ denotes the channel wise concatenation of tensor, $W^O \in \mathbb{R}^{d\times d}$ is the output projection weights, and each head $\operatorname{head}_i$ is calculated as:
\begin{equation}
    \operatorname{head}_{i} = \operatorname{softmax}\left(\frac{Q_iK_i^\top}{\sqrt{d/h}} + B\right)V_i
\end{equation}

where $Q_i = \mathbf{x}'_iW^Q_i, K_i=\mathbf{x}'_iW^K_i, V_i=\mathbf{x}'_iW^V_i \in \mathbb{R}^{M^2\times \frac{d}{h}}$ are the query, key and value tensors, which are created from the local window with $M \times M$ patches with $\frac{d}{h}$ feature dimensions by linearly projecting with learnable weights of $W^Q$, $W^K$, and $W^V \in \mathbb{R}^{\frac{d}{h}\times \frac{d}{h}}$. $B \in \mathbb{R}^{M^2 \times M^2}$ is the relative position bias \cite{liu2021swin} that introduces relative positional information to the model. 

Because the WSA applies the self-attention on the local window, WSA alone cannot obtain a global context of the image. To alleviate this issue, Swin Transformer stacks Transformer blocks using WSA and alternates the window location by half of the window size to gradually build global context by integrating information from different windows. Specifically, the Swin Transformer block consists of a shifted WSA, followed by a 2-layer FFN with GELU activation function in between, which is formulated as:

\begin{equation}
\begin{split}
    \hat{\mathbf{x}}^l &= \mathbf{x}^l + \operatorname{WSA}(norm(\mathbf{x}^l))\\
    \mathbf{x}^{l+1} &= \hat{\mathbf{x}}^l + \operatorname{FFN}(norm(\hat{\mathbf{x}}^l))
\end{split}
\end{equation}

where the $norm$ indicates the LayerNorm \cite{ba2016layer} operation, $\operatorname{FFN}$ indicates the feed-forward network, and partitioning of the input $\mathbf{x}$ is shifted by $(\lfloor \frac{M}{2}, \frac{M}{2} \rfloor )$ from the regularly partitioned windows when layer $l$ is even. This process is illustrated in Figure \ref{fig:enc_block}. For each Transformer Encoder Block, we denote the set of the total number of layers as $\mathcal{L}_s$.

\subsubsection{Patch Merging} In order to generate a hierarchical representation, the spatial resolution of each Transformer Encoder Block is reduced by half through the Patch Merging layer.
The Patch Merging layer takes a feature map size of $\mathbf{x} \in \mathbb{R}^{h \times w \times d}$ as an input. 
The layer first splits and gathers the feature in a checkerboard pattern, creating four sub-feature maps $\mathbf{x}_1$ to $\mathbf{x}_4$ with half of the spatial dimension of the original feature map, where $\mathbf{x}_1$ contains pixels from 'black' squares in even rows, $\mathbf{x}_2$ from 'white' squares in even rows, $\mathbf{x}_3$ from 'black' squares in odd rows, and $\mathbf{x}_4$ from 'white' squares in odd rows. Then these four feature maps are concatenated along the channel dimension, resulting in a tensor of size $h/2 \times w/2 \times 4d$. Finally, the linear projection is applied to reduce the channel dimension from $4d$ to $2d$.

\subsection{CNN Stem}

\begin{figure}[!t]
    \centering
    \includegraphics[width=\linewidth]{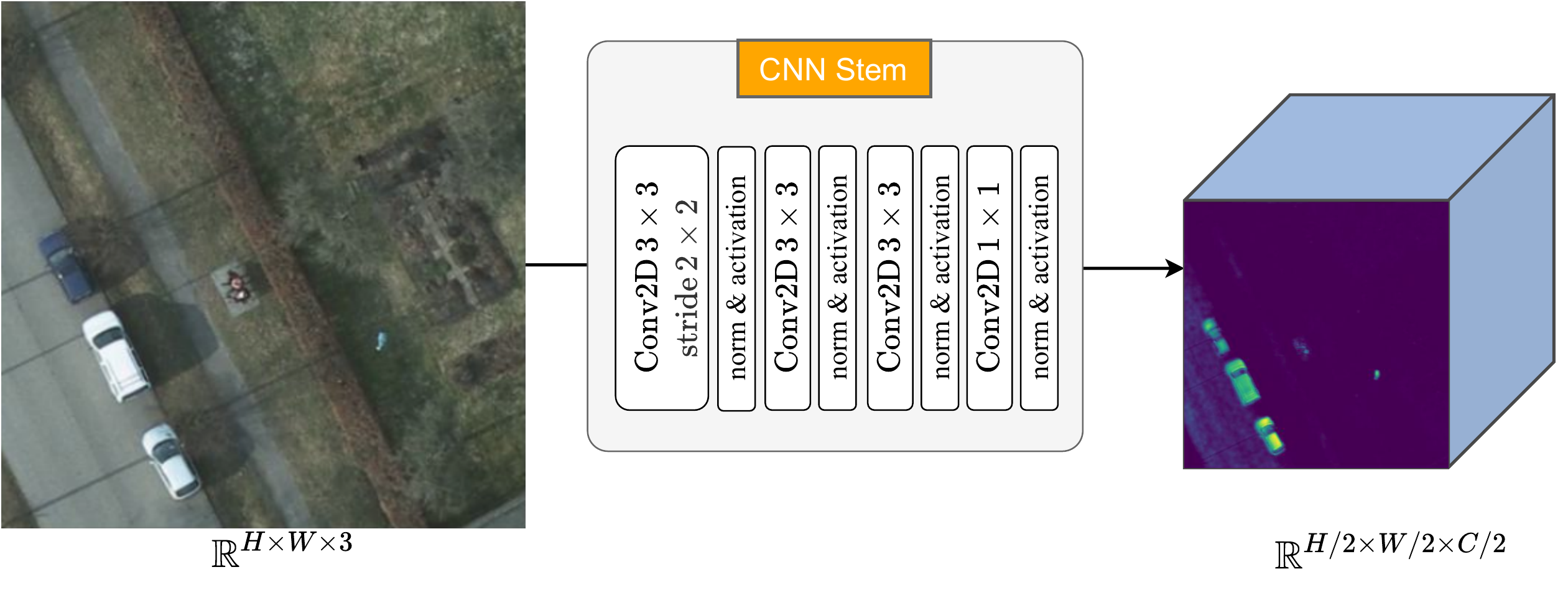}
    \caption{An illustration of the CNN Stem. The Stem takes the input image and produces feature maps with half of the original spacial resolution.}
    \label{fig:stem}
\end{figure}

Although our Transformer encoder is favored in semantic segmentation by using smaller patch size, it may discard the fine-grained details that are especially important in aerial images, which contain tiny and dense objects. To this end, we propose a simple yet effective way to inject the low-level features of the input image to our decoder through a convolutional stem module. 
This module is expected to model the local spatial contexts of images parallel with the patch embedding layer. As shown in the Figure \ref{fig:stem}, our CNN Stem consists of four convolution layers, each followed by BatchNorm \cite{ioffe2015batch} and GELU \cite{hendrycks2016gaussian} activation layers. The first $3\times 3$ convolutional layer with stride of $2\times 2$ reduces the input spacial size into half and through the following three layers of convolution, we obtain local features for tiny and dense objects.

\subsection{Multi-Dilated CNNs Decoder}
While local fine-grained feature is important for segmenting tiny objects, we want to consider the global context at the same time.
In the decoder, we propose to use multiple dilated convolutional operations in parallel with different dilation rates to obtain wider context for decoding without any additional parameters. The Multi-Dilated CNNs Decoder contains a sequence of Multi-Dilated CNNs (MDC) Block followed by Deconvolutional (Deconv) block, which are detailed as follows.

\subsubsection{MDC Block}
A MDC Block is defined by three params [$r_1, r_2, r_3$] corresponding to three receptive fields, and consists of three parts of Pre-Channel Mixer, Dilated Convolutional Layer (DCL) and Post-Channel Mixer.

The MDC Block starts by applying Pre-Channel Mixer to the input, which is the concatenation of previous MDC block's output and the skip connection from the mirrored encoder, in order to exchange the information in channel dimension. The channel mixing operation can be implemented with any operator that enforces the information exchange in channel dimension. Here, Pre-Channel Mixer is implemented as a point-wise convolition layer without any normalization or activation layer.

\begin{figure}
    \centering
    \includegraphics[width=1.05\linewidth]{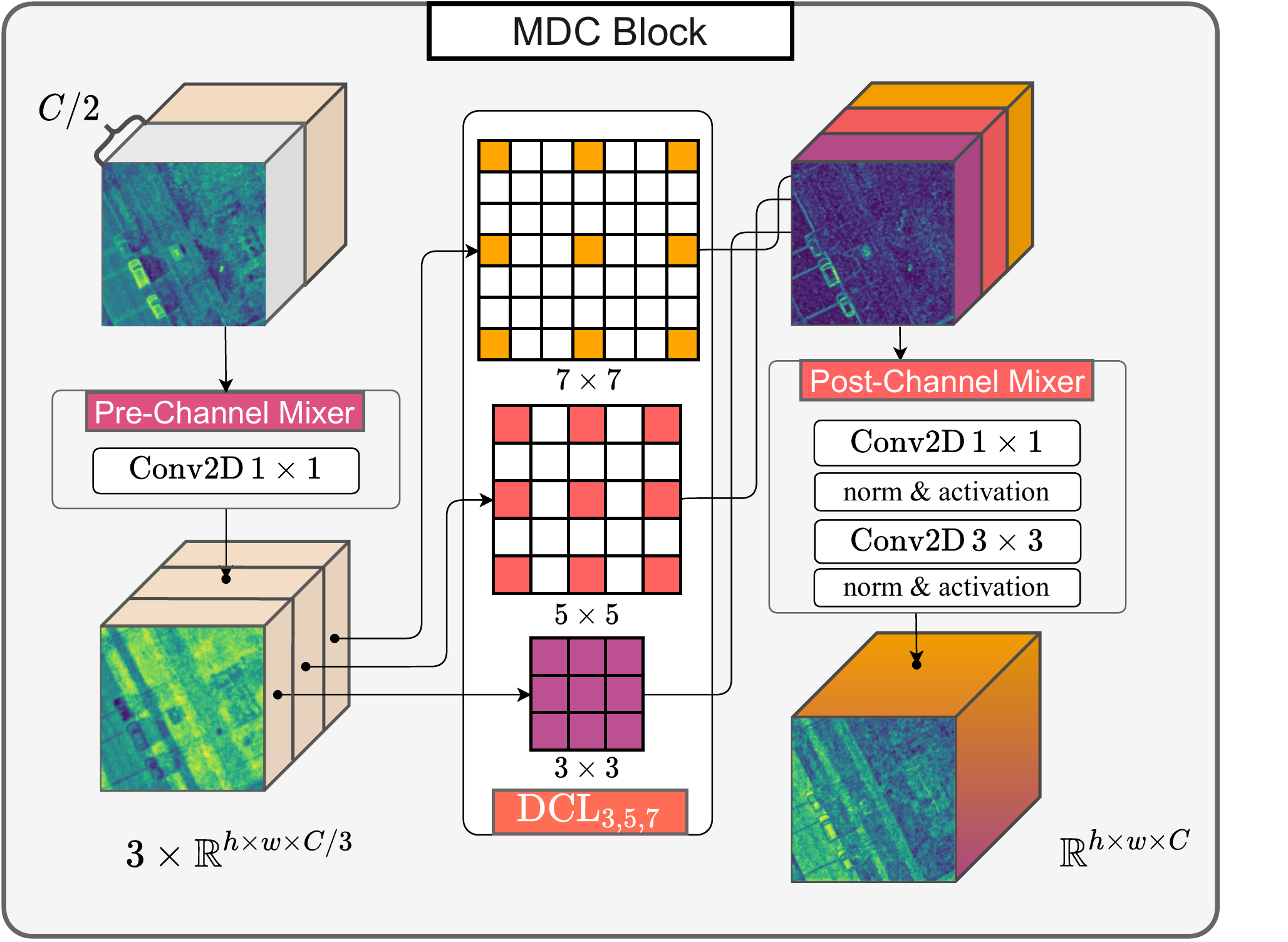}
    \caption{An illustration of the MDC Block, which consists of Pre-Channel Mixer, DCL, and Post-Channel Mixer.}
    \label{fig:mdc}
\end{figure}

The DCL utilizes three convolutional kernels with different dilation rates of $d_1$, $d_2$, and $d_3$, which allows to obtain multi-scale receptive fields.   
We can calculate the length of one side of a receptive field $r$ of dilated convolution given a kernel size $k$ and a dilation rate $d$ as follows:  

\begin{equation}
    r_i = d_i (k - 1) + 1
\end{equation}

where the kernel size $k$ is established as $3$ for receptive fields that exceed $3\times 3$ in size, and as $1$ for those receptive fields that are smaller. 
We will denote the dilated convolutional operation with receptive field of $r \times r$ as $\operatorname{Conv}_r(\cdot)$. Then we can formulate our DCL as follows:

\begin{equation}
    \operatorname{DCL}_{r_1, r_2, r_3}(\mathbf{x}) = [
    \operatorname{Conv}_{r_1}(\mathbf{x}_1); 
    \operatorname{Conv}_{r_2}(\mathbf{x}_2);
    \operatorname{Conv}_{r_3}(\mathbf{x}_3)]
\end{equation}

\begin{table*}[!t]
\renewcommand{\arraystretch}{1.2}
\setlength{\tabcolsep}{1pt}
\newcommand{\isaidscore}[1]{#1}
\newcommand{\paperscore}[1]{#1}
\centering
\caption{Performance comparison on \textbf{iSAID} \textit{valset} between our \model and other SOTA approaches. We report performance on mIoU and IoU for each category. The \textbf{bold} and \underline{\textit{italic-underline}} values in each column show the best and the second best performances.}

\resizebox{\linewidth}{!}{
\begin{tabular}{l|c|c|ccccc|cccc|cccccc}
\toprule
\multirow{3}{*}{\textbf{Method}} & \multirow{3}{*}{\textbf{Year}} &\multirow{3}{*}{\textbf{mIoU} $\uparrow$} & \multicolumn{15}{c}{\textbf{IoU} per category\footnotemark{} $\uparrow$ } \\ \cline{4-18}
 &  & & \multicolumn{5}{c|}{Vehicles} & \multicolumn{4}{c|}{Artifacts} & \multicolumn{6}{c}{Fields}  \\ \cline{4-18}
 &  & & \textbf{LV} & \textbf{SV} & \textbf{Plane} & \textbf{HC} & \textbf{Ship} & \textbf{ST} & \textbf{Bridge} & \textbf{RA} & \textbf{Harbor} & \textbf{BD} & \textbf{TC} & \textbf{GTF} & \textbf{SBF} & \textbf{SP} & \textbf{BC} \\ 
\hline
UNet \cite{ronneberger2015u} & 2015 & $37.4$ & $49.9$ & $35.6$ & $74.7$ & $0.0$ & $49.0$ & $0.0$ & $7.5$ & $46.5$ & $45.6$ & $6.5$ & $78.6$ & $5.5$  & $9.7$ & $38.0$ & $22.9$\\
PSPNet \cite{zhao2017pyramid} & 2017& $60.3$ & $58.0$ & $43.0$ & $79.5$ & $10.9$ & $65.2$ & $52.1$ & $32.5$ & $68.6$ & $54.3$ & $75.7$ & $85.6$ & $60.2$ & $71.9$ & $46.8$ & $61.1$\\
DeepLabV3 \cite{chen2017rethinking} & 2017 & $59.0$ & $54.8$ & $33.7$ & $75.8$ & $31.3$ & $59.7$ & $50.5$ & $32.9$ & $66.0$ & $45.7$ & $77.0$ & $84.2$ & $59.6$ & $72.1$ & $44.7$ & $57.9$ \\
DeepLabV3+ \cite{chen2018encoder}& 2018 & $61.4$ & $61.9$ & $46.7$ & $82.1$ & $0.0$ & $66.2$ & $71.5$ & $37.5$ & $63.1$ & $56.9$ & $73.1$ & $87.2$ & $56.2$ & $73.8$ & $46.6$ & $59.8$ \\

HRNet \cite{sun2019deep} & 2019 & $62.3$ & $61.6$ & $48.5$ & $82.3$ & $6.9$ & $67.5$ & $70.3$ & $38.4$ & $65.7$ & $54.7$ & $75.4$ & $87.1$ & $55.5$ & $75.5$ & $46.4$ & $62.1$\\
FarSeg \cite{zheng2020foreground} & 2020 & 63.7 & 60.6 & 46.3 & 82.0 & 35.8 & 65.4 & 61.8 & 36.7 & 71.4 & 53.9 & 77.7 & 86.4 & 56.7 & 72.5 & 51.2 & 62.1\\

HMANet \cite{niu2021hybrid} & 2021 & \paperscore{62.6} & \paperscore{59.7} & \paperscore{50.3} & \paperscore{83.8} & \paperscore{32.6} & \paperscore{65.4} & \paperscore{70.9} & \paperscore{29.0} & \paperscore{62.9} & \paperscore{51.9} & \paperscore{74.7} & \paperscore{88.7} & \paperscore{54.6} & \paperscore{70.2} & \paperscore{51.4} & \paperscore{60.5}\\
PFNet \cite{li2021pointflow} & 2021 & 66.9 & 64.6 & 50.2 & 85.0 & 37.9 & 70.3  & 74.7 & 45.2 & 71.7 & 59.3 & 77.8 & 87.7 & 59.5 & 75.4 & 50.1  & 62.2\\
Segformer \cite{xie2021segformer}  &2021 & 65.6  & 64.7 &  51.3& 85.1 & 40.3 & 70.8 &  73.9 & 40.8 & 60.9 & 56.9 & 74.6 & 87.9 & 58.9 & 75.0 & 51.2 & 59.1\\

FactSeg \cite{factseg2022} & 2022 & 64.8  & 62.7 & 49.5 & 84.1 & \secondplace{42.7} & 68.3 & 56.8 & 36.3 & 69.4 & 55.7 & 78.4 & 88.9 & 54.6 & 73.6 & 51.5 & 64.9\\
BSNet \cite{hou2022bsnet} & 2022 & \paperscore{63.4} & \paperscore{63.4} & \paperscore{46.6} & \paperscore{81.8} & \paperscore{31.8} & \paperscore{65.3} & \paperscore{69.1} & \paperscore{41.3} & \paperscore{70.0} & \paperscore{57.3} & \paperscore{76.1} & \paperscore{86.8} & \paperscore{50.3} & \paperscore{70.2} & \paperscore{48.8} & \paperscore{55.9}\\
AANet \cite{xue2022aanet} & 2022 & 66.6 & 63.2 & 48.7 & 84.6 & 41.8 & 71.2 & 65.7 & 40.2 & 72.4 & 57.2 & \secondplace{80.5} & 88.8 & \paperscore{60.5} & 73.5 & \secondplace{52.3} & \secondplace{65.4}\\
RSP-Swin-T \cite{wang2022empirical} & 2022 & 64.1 & 62.0 & 50.6 & 85.2 & 37.6 & 67.0 & 74.6 & 44.3 & 64.9 & 53.8 & 73.7 & 70.7 & 60.1 & 76.2 & 46.8 & 59.0\\
Ringmo \cite{sun2022ringmo} & 2022 & \paperscore{67.2} & \paperscore{63.9} & \paperscore{51.2} & \paperscore{85.7} & \paperscore{40.1} & \secondplace{73.5} & \paperscore{73.0} & \paperscore{43.2} & \paperscore{67.3} & \paperscore{58.9} & \paperscore{77.0} & \paperscore{89.1} & \secondplace{63.0} & \textbf{78.5} & \paperscore{48.9} & \paperscore{62.5}\\
RSSFormer \cite{xu2023rssformer} & 2023 & \paperscore{65.9} & --- & --- & --- & --- & --- & --- & --- & --- & --- & --- & --- & --- & --- & --- & ---\\

\hline
\textbf{\model-T} & --& \isaidscore{\iSAIDTiny} & \secondplace{67.0} & 52.6 & 86.1 & 42.0 & 68.6 & \secondplace{74.9} & \secondplace{45.3} & \secondplace{73.0} & 58.2 & 77.5 & 88.8 & 57.5 & 75.1 & 50.5 & 63.4\\
\textbf{\model-S} & --& \secondplace{\iSAIDSmall} & 66.5 & \secondplace{53.6} & \textbf{86.5} & 40.0 & \isaidscore{72.1} & 74.1 & 44.8 & \textbf{74.0}& \textbf{60.9} & 78.8 & \secondplace{89.2} & 59.5 & \isaidscore{77.0} & 52.1 & \textbf{66.5}\\
\textbf{\model-B} & --& \textbf{\iSAIDBase} & \textbf{67.8} & \textbf{53.7} & \textbf{86.5} & \textbf{46.7} & \textbf{75.1} & \textbf{76.3} & \textbf{46.8} & 66.1 & \secondplace{60.8} & \textbf{81.5} & \textbf{89.8} & \textbf{65.0} & \secondplace{78.3} & \textbf{52.4} & 62.4\\
\bottomrule
\end{tabular}
}
\label{tab:iSAID}
\end{table*}

where %
$\mathbf{x} = [\mathbf{x}_1;\mathbf{x}_2;\mathbf{x}_3]$, i.e., the tensor after the Pre-Channel Mixer $\mathbf{x}$ is sliced into three sub-tensors with equivalent channel length.
As we split feature to process with DCL with three different spacial resolution, we applied a Post-Channel Mixer to exchange the information from the three convolutional layers. We implemented the Post-Channel Mixer with sequence of point-wise and $3 \times 3$ convolition layer, each of which are followed by BatchNorm and ReLU activation layers.
This lets us formulate the Multi-Dilated Convolution (MDC) block as follows. The entire operation for MDC Block is illustrated in Figure\ref{fig:mdc}.
\begin{equation}
    \operatorname{MDC}(\mathbf{x}) = \operatorname{PostMixer}(\operatorname{DCL}_{r_1, r_2, r_3}(\operatorname{PreMixer}(\mathbf{x})))
\end{equation}

where $\operatorname{PreMixer}$ refers to the Pre-Channel Mixer and $\operatorname{PostMixer}$ refers to the Post-Channel Mixer. 

\footnotetext{Categories in iSAID dataset: Large Vehicle (LV), Small Vehicle (SV), Plane, Helicopter (HC), Ship, Storage Tank (ST), Bridge, Roundabout (RA), Harbor, Baseball Diamond (BD),  Tennis Court (TC), Ground Track Field (GTF), Soccerball Field (SBF), Swimming Pool (SP), and Basketball Court (BC).}

\subsubsection{Deconv Block} 
The Deconv Block employs the transposed convolution layer, which serves to increase the spatial dimensions of the feature map by a factor of two, while concurrently decreasing the channel dimension by half. We also add the BatchNorm and ReLU activation layers after the transposed convolution operation.

\subsection{Loss Function}
We supervise the network with Cross Entropy Loss, which can be formulated as follows: 

\begin{equation}
    \mathcal{L}_{CE} = - \sum^{n}_{i=1}{t_i\log(p_i)}
\end{equation}

where $t_i$ represents the ground truth and $p_i$ is the softmax probability for the $i^{th}$ class.

\begin{table*}[!t]
\renewcommand{\arraystretch}{1.2}
\setlength{\tabcolsep}{5pt}
\newcommand{\numValues}{3}
\newcommand{\paperscore}[1]{#1}
\newcommand{\potsdamscore}[1]{#1}
\newcommand{\potsdamscoreold}[1]{\score[green]{#1}{\numValues}}

\centering
\caption{Performance comparison on \textbf{Potsdam} \textit{valset with clutter }.
We report performance on mIoU, OA, mF1, and F1 score for each category. Note that both train and evaluation are done on the eroded dataset. The \textbf{bold} and \underline{\textit{italic-underline}} values in each column show the best and the second best performances.}

\resizebox{\textwidth}{!}{
\begin{tabular}{l|c|ccc|cccccc}
\toprule
\textbf{\multirow{2}{*}{Method}}	& \textbf{\multirow{2}{*}{Year}}& \textbf{\multirow{2}{*}{mIoU $\uparrow$}} & \textbf{\multirow{2}{*}{OA $\uparrow$}}&\textbf{\multirow{2}{*}{mF1 $\uparrow$}} & \multicolumn{6}{c}{\textbf{F1} per category\footnotemark{} $\uparrow$ }\\
\cline{6-11}
&&&& & \textbf{Imp. Surf.} & \textbf{Building} & \textbf{Low Veg.} & \textbf{Tree} & \textbf{Car} & \textbf{Clutter}\\
\midrule

FCN \cite{long2015fully} & 2015 & \paperscore{64.2} & --- &\paperscore{75.9} &\paperscore{87.6} &\paperscore{91.6} &\paperscore{77.8} &\paperscore{84.6} &\paperscore{73.5} &\paperscore{40.3}\\
PSPNet \cite{zhao2017pyramid} & 2017 & 77.1 & 90.1 & 85.6 & 92.6 & 96.2 & 86.2 &  88.0  & 95.3 & 55.4\\
DeeplabV3 \cite{chen2017rethinking} & 2017 & 77.2 & 90.0 & 85.6 & 92.4 & 95.9 & 86.4 & 87.6 & 94.9 & 56.7\\

UPerNet \cite{xiao2018unified} & 2018 & 76.8 & 89.7 & 85.6 & 92.5 & 95.5 & 85.5 & 87.5 & 94.9 & 58.0\\
DeepLabV3+ \cite{chen2018encoder} & 2018 & 77.1 & 90.1 & 85.6 & 92.6 & 96.4 & 86.3 & 87.8 & 95.4 & 55.1\\
Denseaspp \cite{yang2018denseaspp} & 2018 & \paperscore{64.7} & --- & \paperscore{76.4} & \paperscore{87.3} &\paperscore{91.1} &\paperscore{76.2} &\paperscore{83.4} &\paperscore{77.1} &\paperscore{43.3}\\

DANet \cite{fu2019dual} & 2019 & \paperscore{65.3} & --- &\paperscore{77.1} &\paperscore{88.5} &\paperscore{92.7} &\paperscore{78.8} &\paperscore{85.7} &\paperscore{73.7} &\paperscore{43.2}\\

EMANet \cite{li2019expectation} & 2019 & \paperscore{65.6} & --- &\paperscore{77.7} &\paperscore{88.2} &\paperscore{92.7} &\paperscore{78.0} &\paperscore{85.7} &\paperscore{72.7} &\paperscore{48.9}\\
CCNet \cite{huang2019ccnet} & 2019 & \paperscore{64.3} & --- &\paperscore{75.9} &\paperscore{88.3} &\paperscore{92.5} &\paperscore{78.8} &\paperscore{85.7} &\paperscore{73.9} &\paperscore{36.3}\\

SCAttNet V2 \cite{li2020scattnet} & 2020 & \paperscore{68.3} &\paperscore{88.0} &\paperscore{78.4} &\paperscore{81.8} &\paperscore{88.8} &\paperscore{72.5} &\paperscore{66.3} &\paperscore{80.3} &\paperscore{20.2}\\
PFNet \cite{li2021pointflow} & 2021 & 75.4 & --- & 84.8 & 91.5 & 95.9 & 85.4 & 86.3 & 91.1 & 58.6\\
Segformer \cite{xie2021segformer} & 2021 & 78.0 & 90.5 & 86.4 & 92.9 & 96.4 & 86.9 & 88.1 & 95.2 & 58.9\\ %

\hline
\textbf{\model-T} & ---& \secondplace{\PotsdamWithClutterTiny} & \potsdamscore{91.1} & \secondplace{87.5} & \textbf{93.5} & \secondplace{96.9} & \potsdamscore{87.2} & \secondplace{89.0} & \secondplace{95.9} & \textbf{62.5}\\
\textbf{\model-S} & ---&\potsdamscore{\PotsdamWithClutterSmall} & \secondplace{91.3} & \potsdamscore{87.2} & \textbf{93.5} & \potsdamscore{97.0} & \secondplace{87.7} & \potsdamscore{88.9} & \textbf{96.0} & \potsdamscore{60.2}\\ %
\textbf{\model-B} & --- &\textbf{\PotsdamWithClutterBase} & \textbf{91.4} & \textbf{87.6} & \textbf{93.5} & \textbf{97.2} & \textbf{88.1} & \textbf{89.3} & \potsdamscore{95.7} & \secondplace{61.9}\\

\bottomrule
\end{tabular}}
\label{tab:potsdam_valset_clutter}
\end{table*} 

\section{Experiments}

\subsection{Datasets}
Our \model is benchmarked on three standard aerial imaging datasets, i.e., iSAID, Potsdam, and LoveDA as below.  

\noindent\textbf{iSAID}: iSAID dataset \cite{Waqas2019} is a large-scale and densely annotated aerial segmentation dataset that contains 655,451 instances of 2,806 high-resolution images for 15 classes (i.e., ship (Ship), storage tank (ST), baseball diamond (BD), tennis court (TC), basketball court (BC), ground field track (GTF), bridge (Bridge), large vehicle (LV), small vehicle (SV), helicopter (HC), swimming pool (SP), roundabout (RA), soccerball field (SBF), plane (Plane), and harbor (Harbor)). This dataset is challenging due to the presence of a large number of objects per image, limited appearance details, a variety of tiny objects, large-scale variations, and high-class imbalance. These images were collected from multiple sensors and platforms with multiple resolutions and image sizes ranging from $800 \times 800$ pixels to  $4000 \times 13,000$ pixels. Follow the experiment setup \cite{zheng2020foreground}, \cite{li2021pointflow}, the dataset is split into 1,411/458/937 images for train/val/test. We train the network on the trainset and benchmark on the valset. Each image is overlap-partitioned into a set of sub-images sized of $896 \times 896$ with a step size of 512 by 512.

\noindent\textbf{Potsdam}: Potsdam dataset \cite{potsdam} contains $38$ high resolution images of $6,000 \times 6,000$ pixels over Potsdam City, Germany, and the ground sampling distance is 5 cm. The dataset is split into 24 images for training and 14 images for validation/testing. There are two modalities included in Potsdam dataset, i.e., true orthophoto (TOP) and digital surface model (DSM). While DSM consists of the near infrared (NIR) band, TOP is corresponding to RGB image. In this work, we use TOP images from Potsdam and ignore DSM images. 
The dataset offers two types of annotations with non-eroded (NE) and eroded (E) options, which respectively with and without the boundary. To avoid ambiguity in labeling boundaries, all experimental results are performed and benchmarked on the eroded boundary dataset. 
Follow experiment setup \cite{he2022swin}, \cite{wang2022empirical} we divide the dataset into 24 images for training and 14 images for testing. The testset of 14 images including 2\_13, 2\_14, 3\_13, 3\_14, 4\_13, 4\_14, 4\_15, 5\_13, 5\_14, 5\_15, 6\_13, 6\_14, 6\_15, and 7\_13. The dataset consists of six categories of surfaces, building, low vegetation, tree, car, and clutter/background. We report the performance in two cases of with and without clutter. Each image is overlap-partitioned into a set of sub-images sized of $512 \times 512$ with a step size of 256 by 256. 
\footnotetext{Categories in Potsdam dataset with Clutter: Impervious Surface (Imp.Surf), Building, Low Vegetation (Low Veg.), Tree, Car, and Clutter/Background. }

\noindent\textbf{LoveDA}: LoveDA dataset \cite{wang2021loveda} consists of 5,987 high resolution images of $1024 \times 1024$ pixels and $30$ cm in spatial resolution. The data include 18 complex urban and rural scenes and 166,768 annotated objects from three different cities (Nanjing, Changzhou, and Wuhan) in China. 
In alignment with the experimental setup delineated in \cite{wang2021loveda}, we partition the dataset into 2,522/1,669/1,796 images for training, validation, and testing, respectively. In evaluation scenarios involving the test set, we amalgamate the training and validation sets of LoveDA to create a combined trainval set, while keeping the test set unaltered.

\subsection{Evaluation Metrics}
To evaluate the performance, we adopt three commonly used metrics: mean intersection over union (mIoU), overall accuracy (OA), and mean F1 score (mF1). 
These metrics are computed based on four fundamental values, namely true positive (TP), true negative (TN), false positive (FP), and false negative (FN). 
The calculation of these four values involves the utilization of the prediction $\text{P}\in \mathbb{R}^{L \times H\times W}$ and class-wise binary groundtruth mask $\text{GT}\in \mathbb{R}^{L \times H\times W}$, where $H$ and $W$ are height and width of the input image and $L$ is the number of classes/categories existing in the input. In the context of multi-class segmentation, these values are computed for each class $l \in [1, 2, ..., L]$ across all pixels.

\begin{table*}[!t]
\caption{Performance comparison on \textbf{Potsdam} \textit{valset without clutter}. We report performance on mIoU, OA, mF1, and F1 score for each category. Note that both train and evaluation are done on the eroded dataset and we ignored the clutter category. The \textbf{bold} and \underline{\textit{italic-underline}} values in each column show the best and the second best performances.}
\label{tab:5potsdam}
\renewcommand{\arraystretch}{1.2}
\setlength{\tabcolsep}{5pt}
\newcommand{\numValues}{3}
\newcommand{\paperscore}[1]{\score{#1}{\numValues}}
\newcommand{\potsdamscore}[1]{\score{#1}{\numValues}}
\newcommand{\potsdamscoreold}[1]{\score[green]{#1}{\numValues}}
\newcommand{\ddagscore}[1]{#1}
\newcommand{\dagscore}[1]{#1}
\centering
\resizebox{\linewidth}{!}{
\begin{tabular}{l|c|ccc|ccccc}
\toprule
\textbf{\multirow{2}{*}{Method}}	&  \textbf{\multirow{2}{*}{Year}} & \textbf{\multirow{2}{*}{mIoU $\uparrow$}} & \textbf{\multirow{2}{*}{OA $\uparrow$}}&\textbf{\multirow{2}{*}{mF1 $\uparrow$}} & \multicolumn{5}{c}{\textbf{F1} per category\footnotemark{} $\uparrow$ }\\
\cline{6-10}
&&&& & \textbf{Imp. Surf.} & \textbf{Building} & \textbf{Low Veg.} & \textbf{Tree} & \textbf{Car}\\
\midrule
DeepLabV3+ \cite{chen2018encoder}& 2018 & \paperscore{81.69} & \paperscore{89.60} & \paperscore{89.79} & \paperscore{92.27} & \paperscore{95.52} & \paperscore{85.71} & \paperscore{86.04} & \paperscore{89.42}\\
DANet \cite{fu2019dual} & 2019 & --- & \dagscore{\potsdamscore{89.72}} & \dagscore{\potsdamscore{89.14}} & \dagscore{\potsdamscore{91.61}} & \dagscore{\potsdamscore{96.44}} & \dagscore{\potsdamscore{86.11}} & \dagscore{\potsdamscore{88.04}} & \dagscore{\potsdamscore{83.54}}\\
LANet \cite{ding2020lanet} &2020 & --- &  \paperscore{90.84} & \paperscore{91.95} & \paperscore{93.05} & \paperscore{97.19} & \paperscore{87.30} & \paperscore{88.04} & \paperscore{94.19}\\
S-RA-FCN \cite{mou2020relation} & 2020& \paperscore{72.5} & \paperscore{88.5} & \paperscore{89.6} & \paperscore{90.7} & \paperscore{94.2} & \paperscore{83.8} & \paperscore{85.8} & \paperscore{93.6}\\
FFPNet \cite{xu2020spatial} & 2020 & \paperscore{86.20} & \paperscore{91.10} & \paperscore{92.44} & \paperscore{93.61} &\paperscore{96.70} &\paperscore{87.31} &\paperscore{88.11} &\paperscore{96.46}\\
ResT \cite{zhang2021rest} & 2021& \ddagscore{\paperscore{85.2}} & \ddagscore{\paperscore{90.6}} & \ddagscore{\paperscore{91.9}} & \ddagscore{\paperscore{92.7}} & \ddagscore{\paperscore{96.1}} & \ddagscore{\paperscore{87.5}} & \ddagscore{\paperscore{88.6}} & \ddagscore{\paperscore{94.8}}\\
ABCNet \cite{li2021abcnet} & 2021&  \paperscore{86.5} & \paperscore{91.3} & \paperscore{92.7} & \paperscore{93.5} & \paperscore{96.9} & \paperscore{87.9} & \paperscore{89.1} & \paperscore{95.8}\\
Segmenter \cite{strudel2021segmenter} &2021 & \paperscore{80.7} & \paperscore{88.7} & \paperscore{89.2} & \paperscore{91.5} & \paperscore{95.3} & \paperscore{85.4} & \paperscore{85.0} & \paperscore{88.5}\\
TransUNet \cite{chen2021transunet} & 2021 & \ddagscore{\paperscore{86.13}} &  --- & \ddagscore{\paperscore{88.09}} & \ddagscore{\paperscore{92.41}} & \ddagscore{\paperscore{94.90}} & \ddagscore{\paperscore{82.89}} & \ddagscore{\paperscore{88.92}} & \ddagscore{\paperscore{91.31}}\\
HMANet \cite{niu2021hybrid} &2021 & \paperscore{87.28} & \paperscore{92.21} & \paperscore{93.20} & \paperscore{93.85} & \paperscore{97.56} & \paperscore{88.65} & \paperscore{89.12} & \paperscore{96.84}\\
DC-Swin \cite{wang2022novel} & 2022 & \potsdamscore{87.56} & \paperscore{92.00} & \paperscore{93.25} & \paperscore{94.19} & \paperscore{97.57} & \paperscore{88.57} & \paperscore{89.62} & \paperscore{96.31}\\
BSNet \cite{hou2022bsnet} & 2022 & \paperscore{77.5} & \paperscore{90.7} & \paperscore{91.5} & \paperscore{92.4} & \paperscore{95.6} & \paperscore{86.8} & \paperscore{88.1} & \paperscore{94.6}\\

UNetFormer \cite{wang2022unetformer} & 2022 & \paperscore{86.8} & \paperscore{91.3}& \paperscore{92.8} &  \paperscore{93.6} & \paperscore{97.2} & \paperscore{87.7} & \paperscore{88.9} & \paperscore{96.5}\\
FT-UNetformer \cite{wang2022unetformer} & 2022 & \potsdamscore{87.5} & \paperscore{92.0} & \paperscore{93.3} & \paperscore{93.9} & \paperscore{97.2} & \paperscore{88.8} & \textbf{89.8} & \paperscore{96.6}\\
UperNet RSP-Swin-T \cite{wang2022empirical} & 2022 & --- & \paperscore{90.78} & \paperscore{90.03} & \paperscore{92.65} & \paperscore{96.35} & \paperscore{86.02} & \paperscore{85.39} & \paperscore{89.75}\\
UperNet-RingMo \cite{sun2022ringmo} & 2022 & --- & \paperscore{91.74} & \paperscore{91.27} &\paperscore{93.63} & \paperscore{97.13} & \paperscore{87.08} & \paperscore{86.36} & \paperscore{92.16} \\

\hline
\textbf{\model-T} & ---  & \potsdamscore{\PotsdamNoClutterTiny} & \potsdamscore{\PotsdamNoClutterTinyOA} & \potsdamscore{\PotsdamNoClutterTinymFscore} & \potsdamscore{95.19} & \secondplace{98.0} & \potsdamscore{89.09} & \potsdamscore{89.11} & \potsdamscore{97.30}\\
\textbf{\model-S} & --- & \secondplace{\PotsdamNoClutterSmall} & \secondplace{93.6} & \secondplace{93.8} & \secondplace{95.3} & \textbf{98.1} & \secondplace{89.2} & \potsdamscore{89.10} & \textbf{97.4} \\
\textbf{\model-B} & --- & \textbf{89.0} & \textbf{93.8} & \textbf{94.0} & \textbf{95.4} & \secondplace{98.0} & \textbf{89.6} & \secondplace{89.7} & \textbf{97.4}\\
\bottomrule
\end{tabular}}
\label{tab:potsdam_valset_noclutter}
\end{table*}

\begin{equation}
\begin{split}
    \text{TP}_l &= \sum_{h=1}^{H}\sum_{w=1}^{W} \text{GT}_{l,h,w} \land \text{P}_{l,h,w} \\
    \text{TN}_l &= \sum_{h=1}^{H}\sum_{w=1}^{W} \lnot (\text{GT}_{l,h,w} \lor \text{P}_{l,h,w}) \\
    \text{FP}_l &= \sum_{h=1}^{H}\sum_{w=1}^{W} \lnot \text{GT}_{l,h,w} \land \text{P}_{l,h,w} \\
    \text{FN}_l &= \sum_{h=1}^{H}\sum_{w=1}^{W} \text{GT}_{l,h,w} \land \lnot \text{P}_{l,h,w}
\end{split}
\end{equation}

\noindent
Based on the four values above, we calculate the IoU, Accuracy (Acc), and F1 of an individual category $l$ as follows:

\begin{equation}
    \text{IoU}_l = \frac{TP_l}{TP_l + FN_l + FP_l}
\end{equation}

\begin{equation}
    \text{Acc}_l = \frac{TP_l + TN_l}{TP_l + TN_l + FN_l + FP_l}
\end{equation}

\begin{equation}
    \text{F1}_l =\frac{2TP_l}{2TP_l + FN_l + FP_l}
\end{equation}

We usually refer $\text{IoU}_l$, $\text{Acc}_l$, and $\text{F1}_l$ as $\text{IoU}$, $\text{Acc}$, $\text{F1}$ of the category $l$.
We the further compute the mIoU, OA, and mF1 as the arithmetic means of the IoU, accuracy, and F1 score, respectively, for each class category.

\begin{equation}
    \text{mIoU} = \frac{1}{L} \sum_{l=1}^{L} \text{IoU}_l
\end{equation}

\begin{equation}
    \text{OA} = \frac{1}{L} \sum_{l=1}^{L} \text{Acc}_l
\end{equation}

\begin{equation}
    \text{mF1} = \frac{1}{L} \sum_{l=1}^{L} \text{F1}_l
\end{equation}

\footnotetext{Categories in Potsdam dataset without Clutter: Impervious Surface (Imp. Surf), Building, Low Vegetation (Low Veg.), Tree, and Car.}

\subsection{Implementation Details}

We trained our \model-T on a single RTX 8000 GPU, and our \model-S and \model-B on two RTX 8000 GPUs. We employed the Adam \cite{kingma2014adam} optimizer with learning rate of $6\times10^{-5}$, weight decay of 0.01, betas of (0.9, 0.999) and batch size of $8$. The experimental models are trained for 160k iterations for LoveDA and Potsdam dataset and 800k iterations for iSAID dataset. During the all training processes, we applied data augmentation such as random horizontal flipping and photometric distortions.

Our \model has been trained on three different backbones, i.e., Swin Transformer-Tiny (Swin-T), Swin Transformer-Small (Swin-B), and Swin Transformer-Base (Swin-B). The first two backbones were pre-trained on Imagenet-1k dataset \cite{deng2009imagenet} and the last backbone was pre-trained on Imagenet-22k dataset \cite{deng2009imagenet}. As a result, we will conduct the experimental performance on three models \model-T, \model-S, and \model-B. As introduced in section \ref{TransEnc}, we delineate the model hyperparameters: 
the number of channels $C$,
window size $M^2$, and
a set of layers $\mathcal{L}= \{\mathcal{L}_s\}_{s=1}^{s=4}$ in Transformer Encoder Blocks, that are specific to each model, as follows:

\begin{itemize}
    \item \model-T: $C=96$, $M^2=7^2$, $\mathcal{L} = \{2,2,6,2\}$ %
    \item \model-S: $C=96$, $M^2=7^2$, $\mathcal{L} = \{2,2,18,2\}$ %
    \item \model-B: $C=128$, $M^2=12^2$, $\mathcal{L} = \{2,2,18,2\}$ %
\end{itemize}

In addition to the aforementioned parameters, we also take note of the receptive field sizes of the MDC Decoder, which remain constant across the models, detailed as follows: $[r_1,r_2,r_3] = \{[1,3,3], [3,3,3], [3,5,7], [3,5,7], [3,5,7]\}$ as demonstrated in Figure \ref{main_fig}.

It is worth highlighting that, relative to the commonly utilized CNN backbones, our model does not significantly increase computational cost, as computational complexities of Swin-T and Swin-S align closely with those of ResNet-50 and ResNet-101, respectively.

\newcommand{\lovedaResults}[8]{#1& #2& #3& #4& #5& #6& #7& #8}

\begin{table*}[!t]
\newcommand{\numValues}{3}
\setlength{\tabcolsep}{5pt}
\newcommand{\paperscore}[1]{\score{#1}{\numValues}}
\newcommand{\lovedascore}[1]{\score{#1}{\numValues}}

\newcommand{\ddagscore}[1]{#1}
\newcommand{\dagscore}[1]{#1}
\caption{Performance comparison on \textbf{LoveDA} testset dataset between our \model and other existing SOTA semantic segmentation approaches. The evaluation is based on a submission to the official server. We report performance on mIoU and IoU for each category. The \textbf{bold} and \underline{\textit{italic-underline}} values in each column show the best and the second best performances.}
\renewcommand{\arraystretch}{1.2}
\centering
\resizebox{\linewidth}{!}{
\begin{tabular}{l|c|c|ccccccc}
\toprule
\textbf{\multirow{2}{*}{Method}}	&  \textbf{\multirow{2}{*}{Year}} & {\multirow{2}{*}{\textbf{mIoU}}} & \multicolumn{7}{c}{\textbf{IoU} per category $\uparrow$}\\
\cline{4-10}
& & & \textbf{Background}& \textbf{Building}& \textbf{Road}& \textbf{Water}& \textbf{Barren}& \textbf{Forest}& \textbf{Agriculture}\\
\midrule

FCN \cite{long2015fully} & 2015 & \ddagscore{\paperscore{46.69}} & \ddagscore{\paperscore{42.60}} & \ddagscore{\paperscore{49.51}} & \ddagscore{\paperscore{48.05}} & \ddagscore{\paperscore{73.09}} & \ddagscore{\paperscore{11.84}} & \ddagscore{\paperscore{43.49}} & \ddagscore{\paperscore{58.30}}\\
UNet \cite{ronneberger2015u} & 2015 & \ddagscore{\paperscore{47.84}} & \ddagscore{\paperscore{43.06}} & \ddagscore{\paperscore{52.74}} & \ddagscore{\paperscore{52.78}} & \ddagscore{\paperscore{73.08}} & \ddagscore{\paperscore{10.33}} & \ddagscore{\paperscore{43.05}} & \ddagscore{\paperscore{59.87}}\\
LinkNet \cite{chaurasia2017linknet} & 2017 & \ddagscore{\paperscore{48.50}} & \ddagscore{\paperscore{43.61}} & \ddagscore{\paperscore{52.07}} & \ddagscore{\paperscore{52.53}} & \ddagscore{\paperscore{76.85}} & \ddagscore{\paperscore{12.16}} & \ddagscore{\paperscore{45.05}} & \ddagscore{\paperscore{57.25}}\\
SegNet \cite{badrinarayanan2017segnet} & 2017 & \paperscore{47.33} & \paperscore{41.75} & \paperscore{51.79} & \paperscore{51.84} & \paperscore{75.37} & \paperscore{10.86} & \paperscore{42.93} & \paperscore{56.74}\\
UNet++ \cite{zhou2018unet++} & 2018 & \ddagscore{\paperscore{48.20}} & \ddagscore{\paperscore{42.85}}  & \ddagscore{\paperscore{52.58}} & \ddagscore{\paperscore{52.82}} & \ddagscore{\paperscore{74.51}} & \ddagscore{\paperscore{11.42}} & \ddagscore{\paperscore{44.42}} & \ddagscore{\paperscore{58.80}}\\
DeeplabV3+ \cite{chen2018encoder} & 2018 & \ddagscore{\paperscore{47.62}} & \ddagscore{\paperscore{43.0}} & \ddagscore{\paperscore{50.9}} & \ddagscore{\paperscore{52.0}} & \ddagscore{\paperscore{74.4}} & \ddagscore{\paperscore{10.4}} & \ddagscore{\paperscore{44.2}} & \ddagscore{\paperscore{58.5}}\\
FarSeg \cite{zheng2020foreground} & 2020 & \paperscore{48.17} & \paperscore{43.39} & \paperscore{51.83} & \paperscore{53.34} & \paperscore{76.07} & \paperscore{10.78} & \paperscore{43.15} & \paperscore{58.62}\\
TransUNet \cite{chen2021transunet} & 2021 & \dagscore{\paperscore{48.9}} & \dagscore{\paperscore{43.0}}& \dagscore{\paperscore{56.1}} &\dagscore{\paperscore{53.7}} &\dagscore{\paperscore{78.0}} &\dagscore{{9.3}} &\dagscore{\paperscore{44.9}} &\dagscore{\paperscore{56.9}}\\
Segmenter \cite{strudel2021segmenter} & 2021 & {\paperscore{47.1}} & \dagscore{\paperscore{38.0}} & \dagscore{\paperscore{50.7}} & \dagscore{\paperscore{48.7}} & \dagscore{\paperscore{77.4}} & \dagscore{\paperscore{13.3}} & \dagscore{\paperscore{43.5}} & \dagscore{\paperscore{58.2}}\\
Segformer \cite{xie2021segformer} & 2021 & \paperscore{49.13051426410675} & \paperscore{42.17773973941803} &\paperscore{56.35729432106018} &\paperscore{50.73705315589905} &\paperscore{78.48021984100342} &\paperscore{17.236937582492828} &\paperscore{45.155069231987} &\paperscore{53.76929044723511}\\
DC-Swin \cite{wang2022novel}& 2022 & \dagscore{\paperscore{50.6}} & \dagscore{\paperscore{41.3}} & \dagscore{\paperscore{54.5}} & \dagscore{\paperscore{56.2}} & \dagscore{\paperscore{78.1}} & \dagscore{\paperscore{14.5}} & \secondplace{47.2} & \dagscore{\paperscore{62.4}}\\

ViTAE-B+RVSA \cite{wang2022advancing}& 2022 & \secondplace{52.4} & --- & --- & --- & --- & --- & --- & ---\\
FactSeg \cite{factseg2022} & 2022 & \dagscore{\paperscore{48.9}} & \dagscore{\paperscore{42.6}} & \dagscore{\paperscore{53.6}} & \dagscore{\paperscore{52.8}} & \dagscore{\paperscore{76.9}} & \dagscore{\paperscore{16.2}} & \dagscore{\paperscore{42.9}} & \dagscore{\paperscore{57.5}}\\
UNetFormer \cite{wang2022unetformer} & 2022 & \secondplace{52.4} & \lovedascore{44.7} & \lovedascore{58.8} &\lovedascore{54.9} & \lovedascore{79.6} & \textbf{20.1} & \lovedascore{46.0} & \lovedascore{62.5}\\
RSSFormer \cite{xu2023rssformer} & 2023 & \secondplace{52.4} & \textbf{52.4} & \textbf{60.7} & \paperscore{55.21} & \paperscore{76.29} & \paperscore{18.73} & \paperscore{45.39} & \paperscore{58.33}\\
\hline
\textbf{\model-T} & --- & \lovedascore{\LovedaTestTiny} & \lovedascore{45.217} & \lovedascore{57.840} & \lovedascore{56.466} & \lovedascore{79.634} & \secondplace{19.2} & \lovedascore{46.121} & \lovedascore{59.535}\\
\textbf{\model-S} & --- & \secondplace{\LovedaTestSmall} & \lovedascore{46.568} & \lovedascore{57.357} & \secondplace{57.3} & \secondplace{80.5} & \lovedascore{15.605} & \lovedascore{46.834} & \secondplace{62.8}\\
\textbf{\model-B} & --- & \textbf{\LovedaTestBase} & \secondplace{47.8} & \textbf{60.7} & \textbf{59.3} & \textbf{81.5} & \lovedascore{17.861194908618927} & \textbf{47.9} & \textbf{64.0}\\

\bottomrule
\end{tabular}
}

\label{tab:LoveDA_Test}
\end{table*}

\subsection{Quantitative Results and Analysis}
The quantitative performance comparisons between our \model with other existing methods are presented in Tables \ref{tab:iSAID}, \ref{tab:potsdam_valset_clutter}, \ref{tab:potsdam_valset_noclutter}, \ref{tab:LoveDA_Test}
for three different datasets under various settings of iSAID (valset), Potsdam (with clutter), Potsdam (without clutter), and LoveDA (testset)
, respectively. For each dataset, we report the performance of the proposed AerialFormer on three backbones of Swin-T, Swin-S, and Swin-B and name them as \model-T, \model-S, and \model-B, respectively. We compare our \model with both CNN-based and Transformer-based image segmentation methods. The comparison on each dataset is detailed as follows:

\subsubsection{\textbf{iSAID semantic segmentation results}}
\label{sec:iSAID_results}
Performance comparisons of our proposed \model with existing state-of-the-art methods on the iSAID dataset are presented in Table \ref{tab:iSAID}. iSAID dataset consists of 15 categories and is divided into three groups of vehicles, artifacts, and fields. In general, we observe that our \model-B achieves the best performance, while both \model-S and \model-T obtain comparable results as the second-best methods. All three models outperform other existing methods significantly. Specifically, our \model-T obtains a mIoU of  $67.5\%$, \model-S achieves a mIoU of  $68.4\%$, and \model-B attains a mIoU of $69.3\%$. Those results present improvements of $0.3\%, 1.2\% \text{ and }2.1\%$  over the previous highest score of $67.2\%$ from RingMo \cite{sun2022ringmo}. Moreover, on some small and dense classes (e.g. small vehicles (SV), planes, helicopters (HC), etc), our \model gains a big margin compared to the existing methods. Take the small vehicles (SV) class as an example, our \model-T achieves $1.4\%$ IoU gain, \model-S gains $2.4\%$ IoU margin, \model-B gains $2.5\%$ IoU margin better than the best existing method i.e., RingMo \cite{sun2022ringmo}. It is worth noting that RingMo utilizes Swin-B as its backbone, which shares a similar computational cost with our \model-B. This analysis further shows that both our \model-T and \model-S, despite being smaller models, outperform the best existing method, RingMo.

\subsubsection{\textbf{Potsdam semantic segmentation results}}
\label{sec:Potsdam_results}
We analyze segmentation performance on Potsdam dataset in two cases of with and without Clutter/Background and the results are summarized in Table \ref{tab:potsdam_valset_clutter} and Table \ref{tab:potsdam_valset_noclutter}, respectively. Clutter class is the most challenging class as it can contain anything except for the five name classes of Impervious Surface, Building, Low Vegetation, Tree, Car. Similar other existing work \cite{li2020scattnet, sun2022ringmo, wang2022unetformer}, we benchmark our \model using various metrics of mIoU, OA, mF1 and F1 per category.

\noindent
\underline{\textbf{Potsdam with Clutter}}:
Table \ref{tab:potsdam_valset_clutter} reports the performance comparisons between our \model with the existing methods on 6 classes (i.e. including Clutter class). It is note that among all existing methods, Segformer \cite{xie2021segformer} is a strong Transformer-based segmentation model and obtains the best performance. Our model gains a remarkable improvement of $1.7\%$ in mIoU, $0.9\%$ in OA, and $1.2\%$ in mF1 compared with the best existing methods Segformer. 

Different from experiment on iSAID (section \ref{sec:iSAID_results}), the tradeoff between performance and model size doesn't seem favorable for this dataset. We speculate that the cause for this could be the difference in the spatial resolution of the datasets. As per \cite{long2021creating}, while the iSAID dataset includes images with spatial resolutions of up to $0.3$ $m$, the spatial resolution of the Potsdam dataset is finer at $0.05$ $m$. Consequently, objects in the Potsdam dataset are represented with more pixels, appearing much larger. This might lessen the requirement for architectural enhancements specifically aimed at improving the segmentation of tiny objects. 

As the most challenging category, F1 score on Clutter is lowest compared to other five categories. Because of the challenging Clutter category, many methods have ignored this category and focused on training the network on only 5 other categories as shown in Table \ref{tab:potsdam_valset_noclutter} below.

\noindent
\underline{\textbf{Potsdam without Clutter}}:
In this experimental setting, the review shows that FT-UNetformer \cite{wang2022unetformer}, HMANet \cite{niu2021hybrid} and DC-Swin \cite{wang2022novel} obtained the best score on mIoU, OA, mF1 metrics and none of them can achieve the best score on all three metrics. In the other hand, our \model-B obtains the best score on all three metrics and gains an improvement of $1.6\%$ mIoU, $1.7\%$ OA, and $0.9\%$ mF1 compared to FT-UNetformer, HMANet, DC-Swin, respectively. Compared to Table \ref{tab:potsdam_valset_clutter} which contains Clutter, we can see that Clutter, when ignored, tends to alleviate the ambiguity amongst the remaining classes.

Similar to the observation on iSAID dataset (section \ref{sec:iSAID_results}), we observe that \model-B achieves the best performance, while both \model-S and \model-T obtain comparable results as the second best methods on Potsdam dataset in both settings of with and without Clutter category.

\subsubsection{\textbf{LoveDA semantic segmentation results}} We report performance comparisons with existing methods on \textit{testset} splits of LoveDA dataset in Table \ref{tab:LoveDA_Test}.
In this experiment, we evaluated our method on the public test server \footnote{https://codalab.lisn.upsaclay.fr/competitions/421} by sending our predictions.  Our smaller model, \model-S, achieves comparable performance to the existing state-of-the-art methods, such as UNetFormer \cite{wang2022unetformer} and RSSFormer \cite{xu2023rssformer}, with an mIoU (mean Intersection over Union) of $52.4\%$. Whereas, our best model, \model-B, shows a significant improvement of $1.7\%$ in mIoU compared to the existing state-of-the-art methods. Notably, \model-B outperforms the existing methods by 4.1\% IoU for the Road category, 5.2\% IoU for the Water category, 2.5\% IoU for the Forest category, and 5.7\% IoU for the Agriculture category. Particularly, 'Road' category, which is typically characterized by narrow and elongated features. Segmenting such objects necessitates both local and global perspectives, a capability our model exhibits effectively.

\subsubsection{\textbf{Network Complexity}}
Besides qualitative analysis, we also include an analysis of the network complexity, as presented in Table \ref{tab:performance}. In this section, we provide details on the model parameters (MB), computation (GFLOPs), and inference time (seconds per image) for our \model, and compare it with two baseline models: Unet \cite{ronneberger2015u}, which is a CNN-based network, and TransUnet \cite{chen2021transunet}, which is a Transformer-based network. To calculate the inference time, we averaged the results of 10,000 runs of the model using $512\times 512$ input with a batch size of 1. While our \model-T has a similar model size and inference time to Unet \cite{ronneberger2015u}, it requires fewer computational resources and achieves significantly higher performance. For example, it achieves a 31.9\% improvement in mIoU on the iSAID dataset. When compared to TransUnet \cite{chen2021transunet}, our \model-T has a comparable inference time of 0.027 seconds per image, as opposed to 0.038 seconds per image. Additionally, it requires a smaller model size, incurs lower computational costs, and achieves higher performance. For instance, it gains a 3.0\% mIoU improvement on the Potsdam validation set without 'Clutter' class, and a 5.2\% mIoU improvement on the LoveDA test set. Even with slightly longer inference times, our models still meet real-time speed requirements. The smallest model in our series, \model-T, can perform inference at a rate of 37 images per second, while \model-S achieves 25.6 images per second. Even the largest model, \model-B, with a model size of 113.82MB, can achieve real-time inference speed at 15.4 images per second.

\begin{table}[h]
\centering
\caption{Performance comparison of our models with different sizes of the backbone.}
\renewcommand{\arraystretch}{1.2}
\resizebox{\linewidth}{!}{
\begin{tabular}{l|c|c|c} 
\toprule
Method & \shortstack{Params\\(MB)} & \shortstack{GFLOPs\\(GB)} & \shortstack{Inference\\ Time (s)}\\
\midrule
Unet \cite{ronneberger2015u} & 29.1 & 203.4 & 0.038 \\ 
TransUnet\cite{chen2021transunet} & 90.7 & 233.7 & 0.023 \\
\hline
\model-T & 42.7 & 49.0 & 0.027\\
\model-S & 64.0 & 72.2 & 0.039\\
\model-B & 113.8 & 126.8 & 0.065\\
\bottomrule
\end{tabular}}

\label{tab:performance}
\end{table}

\begin{figure*}
\centering    
\includegraphics[width=0.95\linewidth]{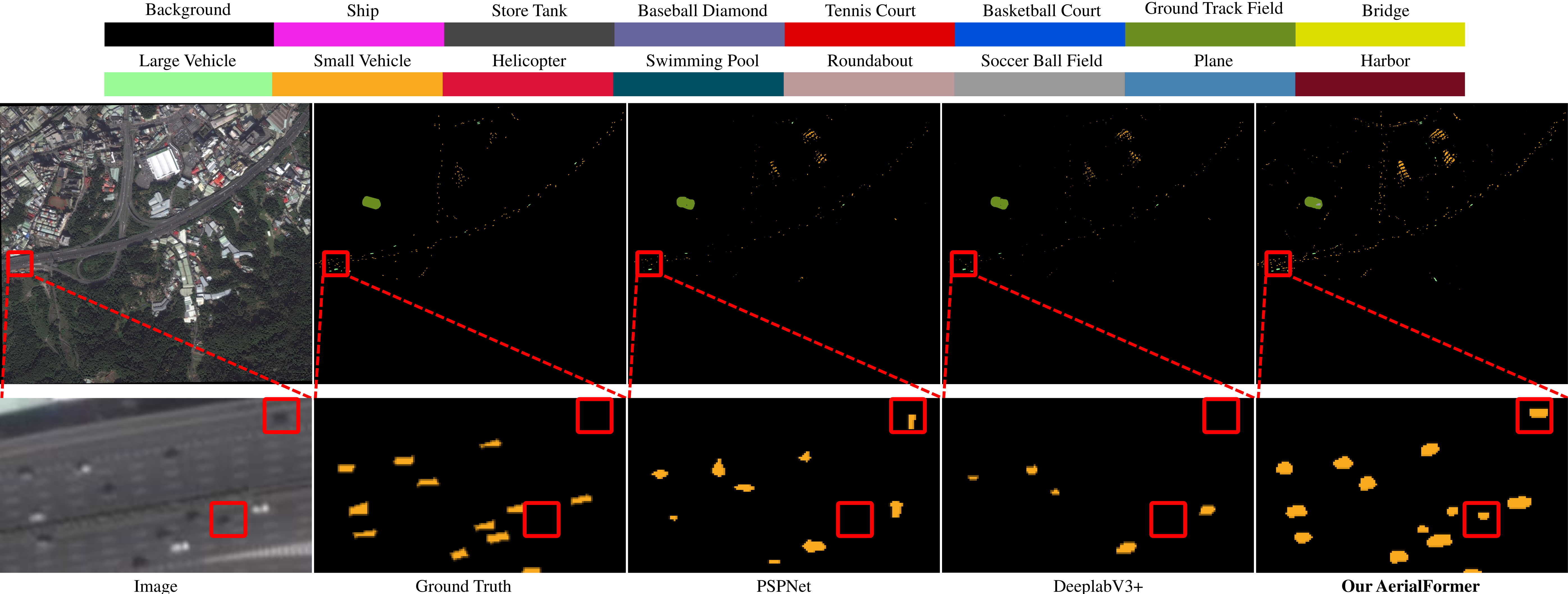}
    \caption{Qualitative comparison between our \model with PSPNet \cite{zhao2017pyramid}, DeepLabV3+ \cite{chen2018encoder} on \textbf{tiny objects}. From left to right are the original image, Groundtruth, PSPNet, DeepLabV3+, and our \model. The first row is overall performance and the second row is zoom-in region. We note that some of the objects that are evident in the input are ignored in the ground truth label.}
    \label{fig:tiny_objects}
\end{figure*}

\begin{figure*}
\centering    
\includegraphics[width=0.95\linewidth]{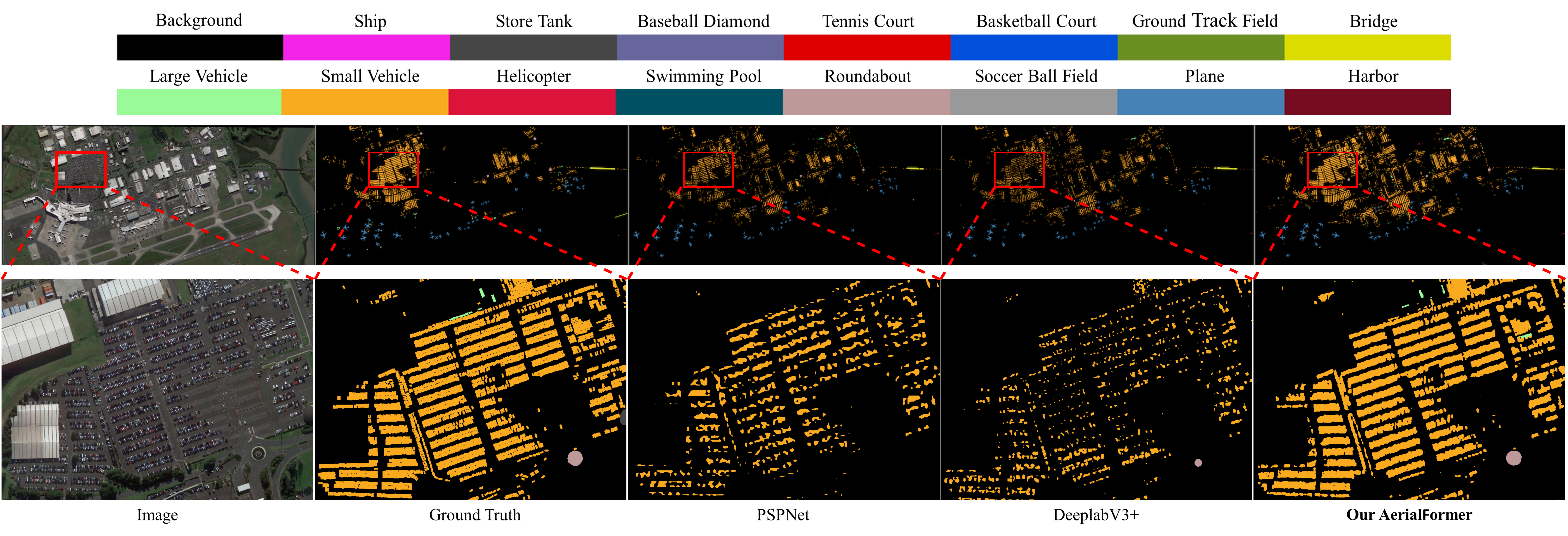}
    \caption{Qualitative comparison between our \model with PSPNet \cite{zhao2017pyramid}, DeepLabV3+ \cite{chen2018encoder} on \textbf{dense objects}. From left to right are the original image, Groundtruth, PSPNet, DeepLabV3+, and our \model. The first row is overall performance and the second row is zoom-in region.}
    \label{fig:dense_objects}
\end{figure*}

\begin{figure*}
\centering    
\includegraphics[width=\linewidth]{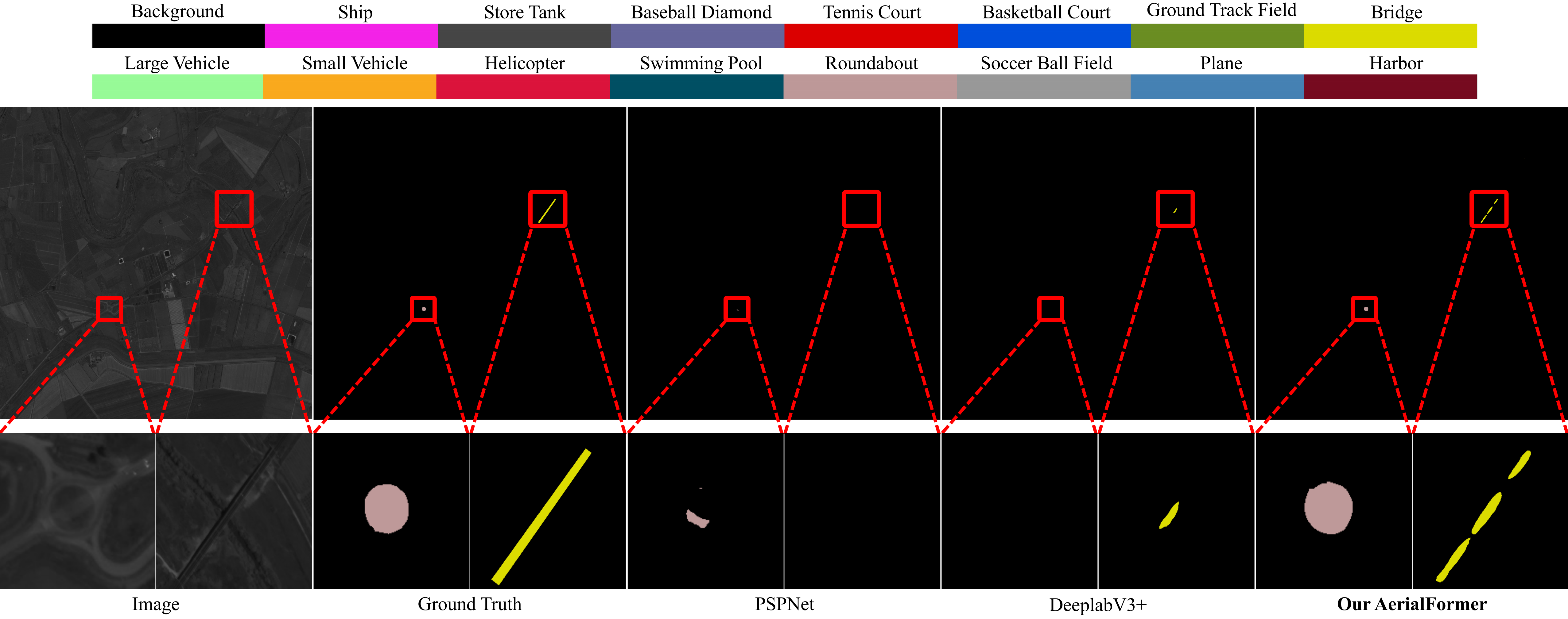}
    \caption{Qualitative comparison between our \model with PSPNet \cite{zhao2017pyramid}, DeepLabV3+ \cite{chen2018encoder} on \textbf{foreground-background imbalance}. From left to right are the original image, Groundtruth, PSPNet, DeepLabV3+, and our \model. The first row is overall performance and the second row is zoom-in region.}
    \label{fig:imbalance}
\end{figure*}

\begin{figure*}
\centering    
\includegraphics[width=0.95\linewidth]{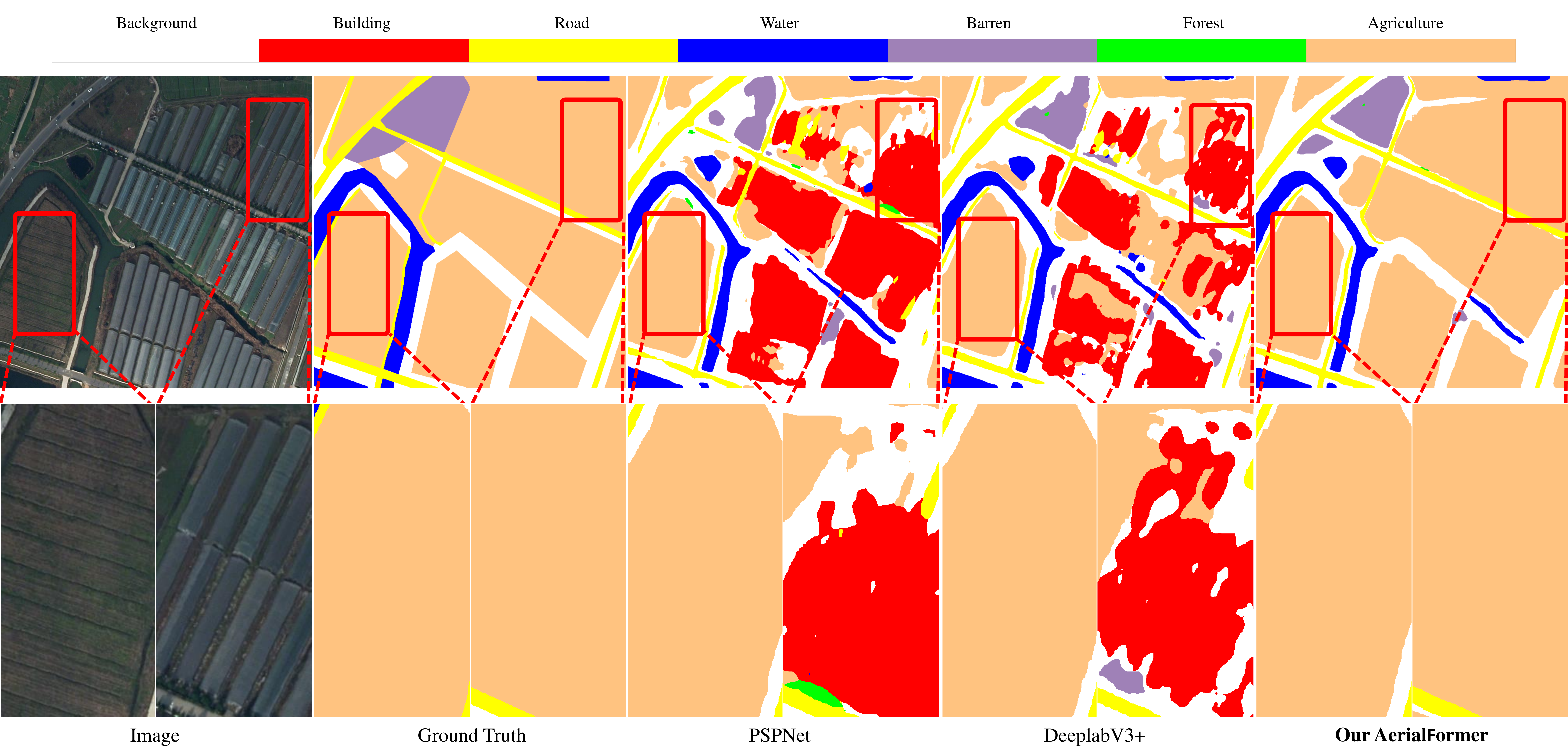}
    \caption{Qualitative comparison between our \model with PSPNet \cite{zhao2017pyramid}, DeepLabV3+ \cite{chen2018encoder} on \textbf{intra-class heterogeneity}: the regions highlighted in the box are both classified under the 'Agriculture' category. However, one region features green lands, while the other depicts greenhouses. From left to right are the original image, Groundtruth, PSPNet, DeepLabV3+, and our \model.}
    \label{fig:intra_class}
\end{figure*}

\begin{figure*}
\centering    
\includegraphics[width=0.95\linewidth]{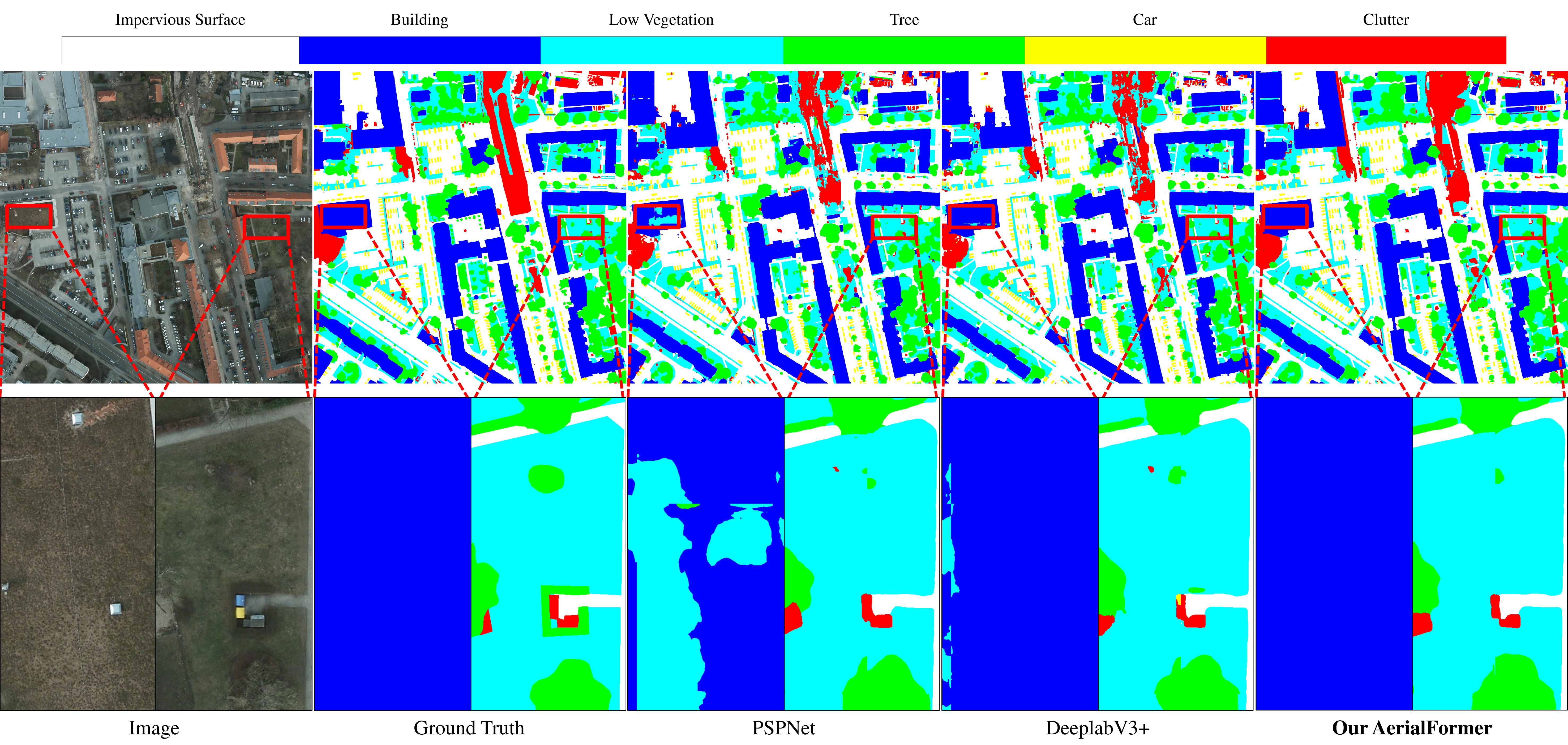}
    \caption{Qualitative comparison between our \model with PSPNet \cite{zhao2017pyramid}, DeepLabV3+ \cite{chen2018encoder} on \textbf{inter-class homogeneity}: the regions highlighted in the box share similar visual characteristics but one region is classified as a 'Building' while the other is classified as belonging to the 'Low Vegetation' category. From left to right are the original image, Groundtruth, PSPNet, DeepLabV3+, and our \model. The first row is overall performance and the second row is zoom-in region.}
    \label{fig:inter_class}
\end{figure*}

\begin{figure*}
    \centering
    \includegraphics[width=0.95\linewidth]{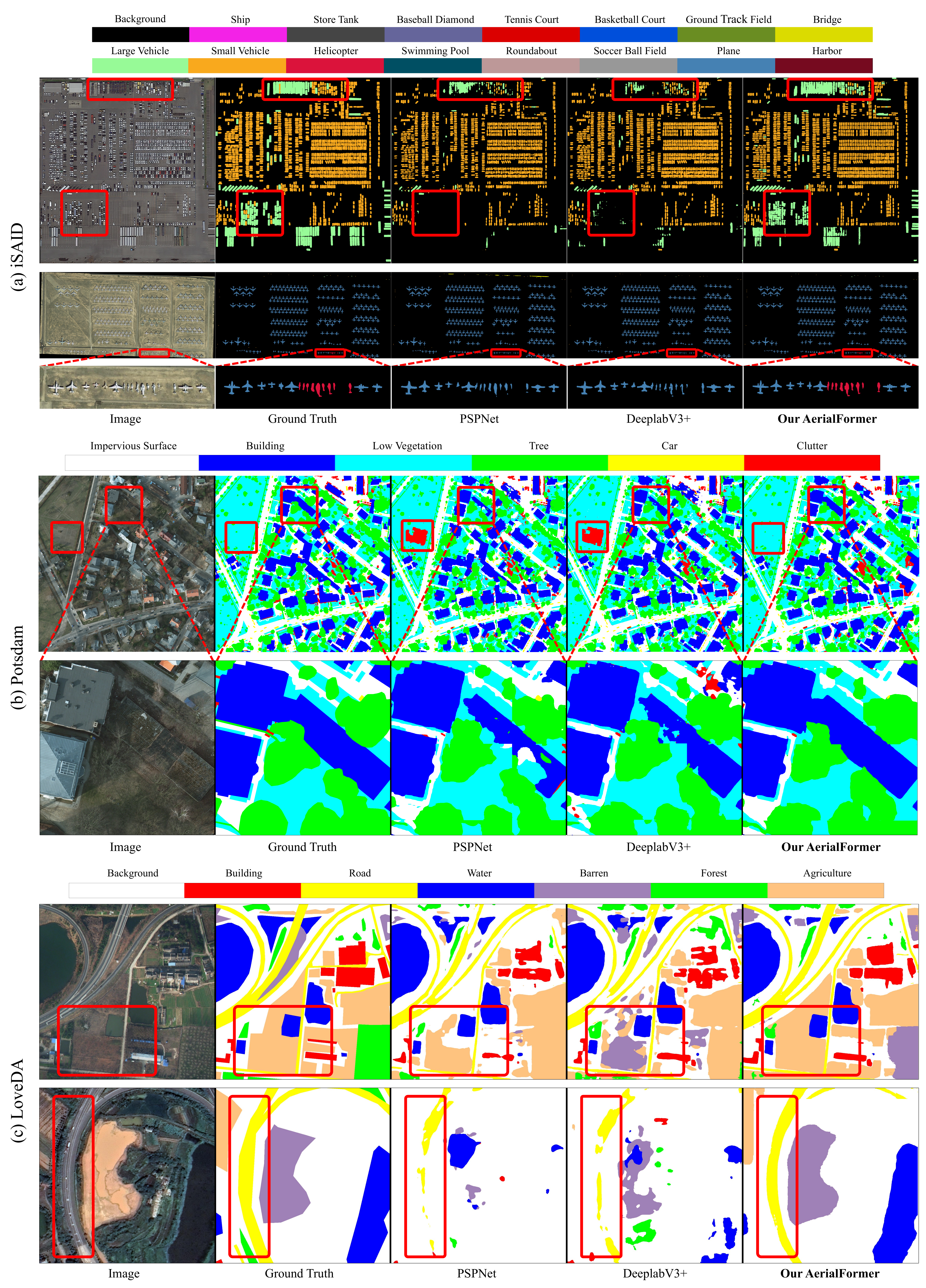}
    \break

    \caption{Qualitative comparison on various datasets: (a) iSAID, (b) Potsdam, and (c) LoveDA. From left to right: original image, ground truth, PSPNet, DeeplabV3+, and our \model. We highlight the major difference in red boxes.}
    \label{fig:qual_all}
\end{figure*}

\subsection{Qualitative Results and Analysis}
We will now present the qualitative results obtained from our model, comparing them
with well-established and robust baseline models, specifically PSPNet \cite{zhao2017pyramid} and DeepLabV3+ \cite{chen2018encoder}. In this section, we will illustrate the advanges of our \model in dealing with challenging characteristics of remote sensing images.

\noindent
\underline{Tiny objects}: 
As evidenced in Fig. \ref{fig:tiny_objects}, our model, \model, is capable of accurately identifying and segmenting tiny objects like cars on the road, which might only be represented by approximately $10 \times 5$ pixels. This showcases the model's remarkable capability to handle small object segmentation in high-resolution aerial images. Additionally, our model demonstrates the ability to accurately segment cars that are not present in the ground truth labels (red boxes). 
However, this poses a problem in evaluating our model, as its prediction could be penalized as false positive even if the prediction is correct based on the given image. 

\noindent
\underline{Dense objects}: 
Fig. \ref{fig:dense_objects} demonstrates the proficient ability of our model in accurately segmenting dense objects, particularly clusters of small vehicles, which often pose challenges for baseline models. The baseline models frequently overlook or struggle with identifying such objects. We ascribe the success of our model in segmenting dense objects to the MDC decoder that can capture the global context and the CNN stem that can bring the local details of the tiny objects.

\noindent
\underline{Foreground-background imbalance}: 
As mentioned in Section \ref{sec:intro}, Introduction, the iSAID dataset exhibits a notable foreground and background imbalance. This imbalance is particularly evident in Fig. \ref{fig:imbalance}, where certain images contain only a few labeled objects. Despite this extreme imbalance, \model demonstrates its capability to accurately segment the objects of interest, as depicted in the figure. 

\noindent
\underline{Intra-class heterogeneity}: 
Fig. \ref{fig:intra_class} visually demonstrates the existence of intra-class heterogeneity in aerial images, where objects of the same category can appear in diverse shapes, textures, colors, scales, and structures. The red boxes indicate two regions that are classified as belonging to the 'Agriculture' category. However, their visual characteristics significantly differ due to the presence of greenhouses. Notably, while baseline models encounter challenges in correctly classifying the region with greenhouses, misclassifying it as 'Building', our proposed model successfully identifies and labels the region as 'Agriculture'. This showcases the superior performance and effectiveness of our model in handling the complexities of intra-class variations in aerial image analysis tasks.

\noindent
\underline{Inter-class heterogeneity}: 
Fig. \ref{fig:inter_class} illustrates the inter-class homogeneity in aerial images, where objects of different classes may exhibit similar visual properties. The regions enclosed within the red boxes represent areas that exhibit similar visual characteristics, i.e., rooftop greened with lawn and the park. However, there is a distinction in the classification of these regions, with the former being labeled as 'Building' and the latter falling into the 'Low Vegetation' category. 
While the baseline models are confused by the appearance and produce mixed prediction, we see our model can produce more robust result. 

\noindent
\underline{Overall performance}: 
Fig. \ref{fig:qual_all} showcases these qualitative outcomes across three datasets: (a) iSAID, (b) Potsdam, and (c) LoveDA. 
Each dataset possesses unique characteristics and presents a wide spectrum of challenges encountered in aerial image segmentation. We highlight the major difference among methods in red boxes. 
Fig. \ref{fig:qual_all} (a) visually demonstrates the efficiency of our model in accurately recognizing \textit{dense and tiny objects}. Unlike the baseline models, which often overlook or misclassify these objects into different categories, our model exhibits its robustness in handling dense and tiny objects, e.g., Small Vehicle (SV) and Helicopter (HC).
As depicted in Fig. \ref{fig:qual_all} (b), our model demonstrates a reduced level of inter-class confusion in comparison to the baseline models. An instance of this is evident in the prediction of building structures, where the baseline models exhibit confusion. In contrast, our model delivers predictions closely aligned with the ground truth.
Similarly, in Fig. \ref{fig:qual_all} (c), our model's predictions are less noisy, further asserting its robustness in scenarios where scenes belong to different categories but exhibit similar visual appearances. As in the quantitative analysis, the performance of our model on the 'Road' class is visually appealing. Our model's ability to accurately delineate road structures, despite their narrow and elongated features, is visibly superior.

\section{Conclusion}
In this study, we have introduced \model, a novel approach specifically designed to address the unique and challenging characteristics encountered in remote sensing image segmentation. These challenges include the presence of tiny objects, dense objects, foreground-background imbalance, intra-class heterogeneity, and inter-class homogeneity. To overcome these challenges, we designed \model by combining the strengths of both Transformers and CNNs architectures, creating a hybrid model that incorporates a Transformer encoder with a multi-dilated CNN decoder. Furthermore, we incorporated a CNN Stem module to facilitate the transmission of low-level, high-resolution features to the decoder. This comprehensive design allows \model to effectively capture global context and local features simultaneously, significantly enhancing its ability to handle the complexities inherent in aerial images.

We have evaluated our proposed \model using three different backbone sizes: Swin Transformer-Tiny, Swin Transformer-Small, and Swin Transformer-Base. Our model was benchmarked on three standard datasets: iSAID, Potsdam, and LoveDA. Through extensive experimentation, we demonstrated that \model-T and \model-S, with smaller model sizes and lower computational costs, achieve performance that is either superior or comparable to existing state-of-the-art methods, ranking them as the second-best performers. Moreover, our proposed \model-B surpasses all existing state-of-the-art methods, showcasing its exceptional performance in the field of remote sensing image segmentation.

\section*{Acknowledgment}
This material is based upon work supported by the National Science Foundation (NSF) under Award No OIA-1946391 RII Track-1, NSF 1920920 RII Track 2 FEC, NSF 2223793 EFRI BRAID, NSF 2119691 AI SUSTEIN, NSF 2236302.

\newpage

\bibliographystyle{plain}

\bibliography{main}

\begin{thebibliography}{100}

\bibitem{potsdam}
2d semantic labeling contest - potsdam.
\newblock {\em International Society for Photogrammetry and Remote Sensing}.

\bibitem{andrade2020evaluation}
RB~Andrade, GAOP Costa, GLA Mota, MX~Ortega, RQ~Feitosa, PJ~Soto, and Christian
  Heipke.
\newblock Evaluation of semantic segmentation methods for deforestation
  detection in the amazon.
\newblock {\em ISPRS Archives; 43, B3}, 43(B3):1497--1505, 2020.

\bibitem{ba2016layer}
Jimmy~Lei Ba, Jamie~Ryan Kiros, and Geoffrey~E Hinton.
\newblock Layer normalization.
\newblock {\em arXiv preprint arXiv:1607.06450}, 2016.

\bibitem{badrinarayanan2017segnet}
Vijay Badrinarayanan, Alex Kendall, and Roberto Cipolla.
\newblock Segnet: A deep convolutional encoder-decoder architecture for image
  segmentation.
\newblock {\em IEEE transactions on pattern analysis and machine intelligence},
  39(12):2481--2495, 2017.

\bibitem{bertasius2016semantic}
Gedas Bertasius, Jianbo Shi, and Lorenzo Torresani.
\newblock Semantic segmentation with boundary neural fields.
\newblock In {\em Proceedings of the IEEE conference on computer vision and
  pattern recognition}, pages 3602--3610, 2016.

\bibitem{carion2020end}
Nicolas Carion, Francisco Massa, Gabriel Synnaeve, Nicolas Usunier, Alexander
  Kirillov, and Sergey Zagoruyko.
\newblock End-to-end object detection with transformers.
\newblock In {\em Computer Vision--ECCV 2020: 16th European Conference,
  Glasgow, UK, August 23--28, 2020, Proceedings, Part I 16}, pages 213--229.
  Springer, 2020.

\bibitem{chaurasia2017linknet}
Abhishek Chaurasia and Eugenio Culurciello.
\newblock Linknet: Exploiting encoder representations for efficient semantic
  segmentation.
\newblock In {\em 2017 IEEE visual communications and image processing (VCIP)},
  pages 1--4. IEEE, 2017.

\bibitem{chen2021transunet}
Jieneng Chen, Yongyi Lu, Qihang Yu, Xiangde Luo, Ehsan Adeli, Yan Wang, Le~Lu,
  Alan~L Yuille, and Yuyin Zhou.
\newblock Transunet: Transformers make strong encoders for medical image
  segmentation.
\newblock {\em arXiv preprint arXiv:2102.04306}, 2021.

\bibitem{chen2017deeplab}
Liang-Chieh Chen, George Papandreou, Iasonas Kokkinos, Kevin Murphy, and Alan~L
  Yuille.
\newblock Deeplab: Semantic image segmentation with deep convolutional nets,
  atrous convolution, and fully connected crfs.
\newblock {\em IEEE transactions on pattern analysis and machine intelligence},
  40(4):834--848, 2017.

\bibitem{chen2017rethinking}
Liang-Chieh Chen, George Papandreou, Florian Schroff, and Hartwig Adam.
\newblock Rethinking atrous convolution for semantic image segmentation.
\newblock {\em arXiv preprint arXiv:1706.05587}, 2017.

\bibitem{chen2018encoder}
Liang-Chieh Chen, Yukun Zhu, George Papandreou, Florian Schroff, and Hartwig
  Adam.
\newblock Encoder-decoder with atrous separable convolution for semantic image
  segmentation.
\newblock In {\em Proceedings of the European conference on computer vision
  (ECCV)}, pages 801--818, 2018.

\bibitem{cheng2022masked}
Bowen Cheng, Ishan Misra, Alexander~G Schwing, Alexander Kirillov, and Rohit
  Girdhar.
\newblock Masked-attention mask transformer for universal image segmentation.
\newblock In {\em Proceedings of the IEEE/CVF Conference on Computer Vision and
  Pattern Recognition}, pages 1290--1299, 2022.

\bibitem{cheng2021per}
Bowen Cheng, Alex Schwing, and Alexander Kirillov.
\newblock Per-pixel classification is not all you need for semantic
  segmentation.
\newblock {\em Advances in Neural Information Processing Systems},
  34:17864--17875, 2021.

\bibitem{dai2017deformable}
Jifeng Dai, Haozhi Qi, Yuwen Xiong, Yi~Li, Guodong Zhang, Han Hu, and Yichen
  Wei.
\newblock Deformable convolutional networks.
\newblock In {\em Proceedings of the IEEE international conference on computer
  vision}, pages 764--773, 2017.

\bibitem{deng2009imagenet}
Jia Deng, Wei Dong, Richard Socher, Li-Jia Li, Kai Li, and Li~Fei-Fei.
\newblock Imagenet: A large-scale hierarchical image database.
\newblock In {\em 2009 IEEE conference on computer vision and pattern
  recognition}, pages 248--255. Ieee, 2009.

\bibitem{ding2019boundary}
Henghui Ding, Xudong Jiang, Ai~Qun Liu, Nadia~Magnenat Thalmann, and Gang Wang.
\newblock Boundary-aware feature propagation for scene segmentation.
\newblock In {\em Proceedings of the IEEE/CVF International Conference on
  Computer Vision}, pages 6819--6829, 2019.

\bibitem{ding2020lanet}
Lei Ding, Hao Tang, and Lorenzo Bruzzone.
\newblock Lanet: Local attention embedding to improve the semantic segmentation
  of remote sensing images.
\newblock {\em IEEE Transactions on Geoscience and Remote Sensing},
  59(1):426--435, 2020.

\bibitem{dosovitskiy2020image}
Alexey Dosovitskiy, Lucas Beyer, Alexander Kolesnikov, Dirk Weissenborn,
  Xiaohua Zhai, Thomas Unterthiner, Mostafa Dehghani, Matthias Minderer, Georg
  Heigold, Sylvain Gelly, et~al.
\newblock An image is worth 16x16 words: Transformers for image recognition at
  scale.
\newblock In {\em International Conference on Learning Representations}, 2021.

\bibitem{fu2019dual}
Jun Fu, Jing Liu, Haijie Tian, Yong Li, Yongjun Bao, Zhiwei Fang, and Hanqing
  Lu.
\newblock Dual attention network for scene segmentation.
\newblock In {\em Proceedings of the IEEE/CVF conference on computer vision and
  pattern recognition}, pages 3146--3154, 2019.

\bibitem{griffiths2019improving}
David Griffiths and Jan Boehm.
\newblock Improving public data for building segmentation from convolutional
  neural networks (cnns) for fused airborne lidar and image data using active
  contours.
\newblock {\em ISPRS Journal of Photogrammetry and Remote Sensing}, 154:70--83,
  2019.

\bibitem{harley2017segmentation}
Adam~W Harley, Konstantinos~G Derpanis, and Iasonas Kokkinos.
\newblock Segmentation-aware convolutional networks using local attention
  masks.
\newblock In {\em Proceedings of the IEEE International Conference on Computer
  Vision}, pages 5038--5047, 2017.

\bibitem{he2019dynamic}
Junjun He, Zhongying Deng, and Yu~Qiao.
\newblock Dynamic multi-scale filters for semantic segmentation.
\newblock In {\em Proceedings of the IEEE/CVF International Conference on
  Computer Vision}, pages 3562--3572, 2019.

\bibitem{he2019adaptive}
Junjun He, Zhongying Deng, Lei Zhou, Yali Wang, and Yu~Qiao.
\newblock Adaptive pyramid context network for semantic segmentation.
\newblock In {\em Proceedings of the IEEE/CVF Conference on Computer Vision and
  Pattern Recognition}, pages 7519--7528, 2019.

\bibitem{he2016deep}
Kaiming He, Xiangyu Zhang, Shaoqing Ren, and Jian Sun.
\newblock Deep residual learning for image recognition.
\newblock In {\em Proceedings of the IEEE conference on computer vision and
  pattern recognition}, pages 770--778, 2016.

\bibitem{he2022swin}
Xin He, Yong Zhou, Jiaqi Zhao, Di~Zhang, Rui Yao, and Yong Xue.
\newblock Swin transformer embedding unet for remote sensing image semantic
  segmentation.
\newblock {\em IEEE Transactions on Geoscience and Remote Sensing}, 60:1--15,
  2022.

\bibitem{hendrycks2016gaussian}
Dan Hendrycks and Kevin Gimpel.
\newblock Gaussian error linear units (gelus).
\newblock {\em arXiv preprint arXiv:1606.08415}, 2016.

\bibitem{hoang2022dam}
Dinh-Hieu Hoang, Gia-Han Diep, Minh-Triet Tran, and Ngan T~H Le.
\newblock Dam-al: Dilated attention mechanism with attention loss for 3d infant
  brain image segmentation.
\newblock In {\em Proceedings of the 37th ACM/SIGAPP Symposium on Applied
  Computing}, pages 660--668, 2022.

\bibitem{hou2022bsnet}
Jianlong Hou, Zhi Guo, Youming Wu, Wenhui Diao, and Tao Xu.
\newblock Bsnet: Dynamic hybrid gradient convolution based boundary-sensitive
  network for remote sensing image segmentation.
\newblock {\em IEEE Transactions on Geoscience and Remote Sensing}, 60:1--22,
  2022.

\bibitem{howard2017mobilenets}
Andrew~G Howard, Menglong Zhu, Bo~Chen, Dmitry Kalenichenko, Weijun Wang,
  Tobias Weyand, Marco Andreetto, and Hartwig Adam.
\newblock Mobilenets: Efficient convolutional neural networks for mobile vision
  applications.
\newblock {\em arXiv preprint arXiv:1704.04861}, 2017.

\bibitem{hsiao2021specialize}
Chi-Wei Hsiao, Cheng Sun, Hwann-Tzong Chen, and Min Sun.
\newblock Specialize and fuse: Pyramidal output representation for semantic
  segmentation.
\newblock In {\em Proceedings of the IEEE/CVF International Conference on
  Computer Vision}, pages 7137--7146, 2021.

\bibitem{hu2020class}
Hanzhe Hu, Deyi Ji, Weihao Gan, Shuai Bai, Wei Wu, and Junjie Yan.
\newblock Class-wise dynamic graph convolution for semantic segmentation.
\newblock In {\em Computer Vision--ECCV 2020: 16th European Conference,
  Glasgow, UK, August 23--28, 2020, Proceedings, Part XVII 16}, pages 1--17.
  Springer, 2020.

\bibitem{hu2018squeeze}
Jie Hu, Li~Shen, and Gang Sun.
\newblock Squeeze-and-excitation networks.
\newblock In {\em Proceedings of the IEEE conference on computer vision and
  pattern recognition}, pages 7132--7141, 2018.

\bibitem{huang2019ccnet}
Zilong Huang, Xinggang Wang, Lichao Huang, Chang Huang, Yunchao Wei, and Wenyu
  Liu.
\newblock Ccnet: Criss-cross attention for semantic segmentation.
\newblock In {\em Proceedings of the IEEE/CVF international conference on
  computer vision}, pages 603--612, 2019.

\bibitem{ioffe2015batch}
Sergey Ioffe and Christian Szegedy.
\newblock Batch normalization: Accelerating deep network training by reducing
  internal covariate shift.
\newblock In {\em International conference on machine learning}, pages
  448--456. pmlr, 2015.

\bibitem{jin2021mining}
Zhenchao Jin, Tao Gong, Dongdong Yu, Qi~Chu, Jian Wang, Changhu Wang, and Jie
  Shao.
\newblock Mining contextual information beyond image for semantic segmentation.
\newblock In {\em Proceedings of the IEEE/CVF International Conference on
  Computer Vision}, pages 7231--7241, 2021.

\bibitem{jin2021isnet}
Zhenchao Jin, Bin Liu, Qi~Chu, and Nenghai Yu.
\newblock Isnet: Integrate image-level and semantic-level context for semantic
  segmentation.
\newblock In {\em Proceedings of the IEEE/CVF International Conference on
  Computer Vision}, pages 7189--7198, 2021.

\bibitem{kingma2014adam}
Diederik~P Kingma and Jimmy Ba.
\newblock Adam: A method for stochastic optimization.
\newblock {\em arXiv preprint arXiv:1412.6980}, 2014.

\bibitem{le2021narrow}
Ngan Le, Toan Bui, Viet-Khoa Vo-Ho, Kashu Yamazaki, and Khoa Luu.
\newblock Narrow band active contour attention model for medical segmentation.
\newblock {\em Diagnostics}, 11(8):1393, 2021.

\bibitem{le2021offset}
Ngan Le, Trung Le, Kashu Yamazaki, Toan Bui, Khoa Luu, and Marios Savides.
\newblock Offset curves loss for imbalanced problem in medical segmentation.
\newblock In {\em 2020 25th International Conference on Pattern Recognition
  (ICPR)}, pages 9189--9195. IEEE, 2021.

\bibitem{le2021multi}
Ngan Le, Kashu Yamazaki, Kha~Gia Quach, Dat Truong, and Marios Savvides.
\newblock A multi-task contextual atrous residual network for brain tumor
  detection \& segmentation.
\newblock In {\em 2020 25th International Conference on Pattern Recognition
  (ICPR)}, pages 5943--5950. IEEE, 2021.

\bibitem{le2018deep}
T~Hoang~Ngan Le, Chi~Nhan Duong, Ligong Han, Khoa Luu, Kha~Gia Quach, and
  Marios Savvides.
\newblock Deep contextual recurrent residual networks for scene labeling.
\newblock {\em Pattern Recognition}, 80:32--41, 2018.

\bibitem{li2022dn}
Feng Li, Hao Zhang, Shilong Liu, Jian Guo, Lionel~M Ni, and Lei Zhang.
\newblock Dn-detr: Accelerate detr training by introducing query denoising.
\newblock In {\em Proceedings of the IEEE/CVF Conference on Computer Vision and
  Pattern Recognition}, pages 13619--13627, 2022.

\bibitem{li2020scattnet}
Haifeng Li, Kaijian Qiu, Li~Chen, Xiaoming Mei, Liang Hong, and Chao Tao.
\newblock Scattnet: Semantic segmentation network with spatial and channel
  attention mechanism for high-resolution remote sensing images.
\newblock {\em IEEE Geoscience and Remote Sensing Letters}, 18(5):905--909,
  2020.

\bibitem{li2018pyramid}
Hanchao Li, Pengfei Xiong, Jie An, and Lingxue Wang.
\newblock Pyramid attention network for semantic segmentation.
\newblock {\em arXiv preprint arXiv:1805.10180}, 2018.

\bibitem{li2021abcnet}
Rui Li, Shunyi Zheng, Ce~Zhang, Chenxi Duan, Libo Wang, and Peter~M Atkinson.
\newblock Abcnet: Attentive bilateral contextual network for efficient semantic
  segmentation of fine-resolution remotely sensed imagery.
\newblock {\em ISPRS Journal of Photogrammetry and Remote Sensing}, 181:84--98,
  2021.

\bibitem{li2019expectation}
Xia Li, Zhisheng Zhong, Jianlong Wu, Yibo Yang, Zhouchen Lin, and Hong Liu.
\newblock Expectation-maximization attention networks for semantic
  segmentation.
\newblock In {\em Proceedings of the IEEE/CVF International Conference on
  Computer Vision}, pages 9167--9176, 2019.

\bibitem{li2021pointflow}
Xiangtai Li, Hao He, Xia Li, Duo Li, Guangliang Cheng, Jianping Shi, Lubin
  Weng, Yunhai Tong, and Zhouchen Lin.
\newblock Pointflow: Flowing semantics through points for aerial image
  segmentation.
\newblock In {\em Proceedings of the IEEE/CVF Conference on Computer Vision and
  Pattern Recognition}, pages 4217--4226, 2021.

\bibitem{li2020improving}
Xiangtai Li, Xia Li, Li~Zhang, Guangliang Cheng, Jianping Shi, Zhouchen Lin,
  Shaohua Tan, and Yunhai Tong.
\newblock Improving semantic segmentation via decoupled body and edge
  supervision.
\newblock In {\em Computer Vision--ECCV 2020: 16th European Conference,
  Glasgow, UK, August 23--28, 2020, Proceedings, Part XVII 16}, pages 435--452.
  Springer, 2020.

\bibitem{liu2022dab}
Shilong Liu, Feng Li, Hao Zhang, Xiao Yang, Xianbiao Qi, Hang Su, Jun Zhu, and
  Lei Zhang.
\newblock Dab-detr: Dynamic anchor boxes are better queries for detr.
\newblock {\em arXiv preprint arXiv:2201.12329}, 2022.

\bibitem{liu2021swin}
Ze~Liu, Yutong Lin, Yue Cao, Han Hu, Yixuan Wei, Zheng Zhang, Stephen Lin, and
  Baining Guo.
\newblock Swin transformer: Hierarchical vision transformer using shifted
  windows.
\newblock In {\em Proceedings of the IEEE/CVF international conference on
  computer vision}, pages 10012--10022, 2021.

\bibitem{long2015fully}
Jonathan Long, Evan Shelhamer, and Trevor Darrell.
\newblock Fully convolutional networks for semantic segmentation.
\newblock In {\em Proceedings of the IEEE conference on computer vision and
  pattern recognition}, pages 3431--3440, 2015.

\bibitem{long2021creating}
Yang Long, Gui-Song Xia, Shengyang Li, Wen Yang, Michael~Ying Yang, Xiao~Xiang
  Zhu, Liangpei Zhang, and Deren Li.
\newblock On creating benchmark dataset for aerial image interpretation:
  Reviews, guidances, and million-aid.
\newblock {\em IEEE Journal of selected topics in applied earth observations
  and remote sensing}, 14:4205--4230, 2021.

\bibitem{factseg2022}
Ailong Ma, Junjue Wang, Yanfei Zhong, and Zhuo Zheng.
\newblock Factseg: Foreground activation-driven small object semantic
  segmentation in large-scale remote sensing imagery.
\newblock {\em IEEE Transactions on Geoscience and Remote Sensing}, 60:1--16,
  2022.

\bibitem{marcos2018land}
Diego Marcos, Michele Volpi, Benjamin Kellenberger, and Devis Tuia.
\newblock Land cover mapping at very high resolution with rotation equivariant
  cnns: Towards small yet accurate models.
\newblock {\em ISPRS journal of photogrammetry and remote sensing},
  145:96--107, 2018.

\bibitem{minaee2021image}
Shervin Minaee, Yuri~Y Boykov, Fatih Porikli, Antonio~J Plaza, Nasser
  Kehtarnavaz, and Demetri Terzopoulos.
\newblock Image segmentation using deep learning: A survey.
\newblock {\em IEEE transactions on pattern analysis and machine intelligence},
  2021.

\bibitem{mou2020relation}
Lichao Mou, Yuansheng Hua, and Xiao~Xiang Zhu.
\newblock Relation matters: Relational context-aware fully convolutional
  network for semantic segmentation of high-resolution aerial images.
\newblock {\em IEEE Transactions on Geoscience and Remote Sensing},
  58(11):7557--7569, 2020.

\bibitem{niu2021hybrid}
Ruigang Niu, Xian Sun, Yu~Tian, Wenhui Diao, Kaiqiang Chen, and Kun Fu.
\newblock Hybrid multiple attention network for semantic segmentation in aerial
  images.
\newblock {\em IEEE Transactions on Geoscience and Remote Sensing}, 60:1--18,
  2021.

\bibitem{o2013use}
Saffron~J O’neill, Maxwell Boykoff, Simon Niemeyer, and Sophie~A Day.
\newblock On the use of imagery for climate change engagement.
\newblock {\em Global environmental change}, 23(2):413--421, 2013.

\bibitem{ronneberger2015u}
Olaf Ronneberger, Philipp Fischer, and Thomas Brox.
\newblock U-net: Convolutional networks for biomedical image segmentation.
\newblock In {\em Medical Image Computing and Computer-Assisted
  Intervention--MICCAI 2015: 18th International Conference, Munich, Germany,
  October 5-9, 2015, Proceedings, Part III 18}, pages 234--241. Springer, 2015.

\bibitem{samie2020examining}
Abdus Samie, Azhar Abbas, Muhammad~Masood Azeem, Sidra Hamid, Muhammad~Amjed
  Iqbal, Shaikh~Shamim Hasan, and Xiangzheng Deng.
\newblock Examining the impacts of future land use/land cover changes on
  climate in punjab province, pakistan: implications for environmental
  sustainability and economic growth.
\newblock {\em Environmental Science and Pollution Research}, 27:25415--25433,
  2020.

\bibitem{schumann2018assisting}
Guy~JP Schumann, G~Robert Brakenridge, Albert~J Kettner, Rashid Kashif, and
  Emily Niebuhr.
\newblock Assisting flood disaster response with earth observation data and
  products: A critical assessment.
\newblock {\em Remote Sensing}, 10(8):1230, 2018.

\bibitem{shafique2022deep}
Ayesha Shafique, Guo Cao, Zia Khan, Muhammad Asad, and Muhammad Aslam.
\newblock Deep learning-based change detection in remote sensing images: A
  review.
\newblock {\em Remote Sensing}, 14(4):871, 2022.

\bibitem{shamsolmoali2020road}
Pourya Shamsolmoali, Masoumeh Zareapoor, Huiyu Zhou, Ruili Wang, and Jie Yang.
\newblock Road segmentation for remote sensing images using adversarial spatial
  pyramid networks.
\newblock {\em IEEE Transactions on Geoscience and Remote Sensing},
  59(6):4673--4688, 2020.

\bibitem{simonyan2014very}
Karen Simonyan and Andrew Zisserman.
\newblock Very deep convolutional networks for large-scale image recognition.
\newblock {\em arXiv preprint arXiv:1409.1556}, 2014.

\bibitem{strudel2021segmenter}
Robin Strudel, Ricardo Garcia, Ivan Laptev, and Cordelia Schmid.
\newblock Segmenter: Transformer for semantic segmentation.
\newblock In {\em Proceedings of the IEEE/CVF international conference on
  computer vision}, pages 7262--7272, 2021.

\bibitem{sun2020mining}
Guolei Sun, Wenguan Wang, Jifeng Dai, and Luc Van~Gool.
\newblock Mining cross-image semantics for weakly supervised semantic
  segmentation.
\newblock In {\em Computer Vision--ECCV 2020: 16th European Conference,
  Glasgow, UK, August 23--28, 2020, Proceedings, Part II 16}, pages 347--365.
  Springer, 2020.

\bibitem{sun2019deep}
Ke~Sun, Bin Xiao, Dong Liu, and Jingdong Wang.
\newblock Deep high-resolution representation learning for human pose
  estimation.
\newblock In {\em Proceedings of the IEEE/CVF conference on computer vision and
  pattern recognition}, pages 5693--5703, 2019.

\bibitem{sun2021sparse}
Peize Sun, Rufeng Zhang, Yi~Jiang, Tao Kong, Chenfeng Xu, Wei Zhan, Masayoshi
  Tomizuka, Lei Li, Zehuan Yuan, Changhu Wang, et~al.
\newblock Sparse r-cnn: End-to-end object detection with learnable proposals.
\newblock In {\em Proceedings of the IEEE/CVF conference on computer vision and
  pattern recognition}, pages 14454--14463, 2021.

\bibitem{sun2022ringmo}
Xian Sun, Peijin Wang, Wanxuan Lu, Zicong Zhu, Xiaonan Lu, Qibin He, Junxi Li,
  Xuee Rong, Zhujun Yang, Hao Chang, et~al.
\newblock Ringmo: A remote sensing foundation model with masked image modeling.
\newblock {\em IEEE Transactions on Geoscience and Remote Sensing}, 2022.

\bibitem{szegedy2017inception}
Christian Szegedy, Sergey Ioffe, Vincent Vanhoucke, and Alexander Alemi.
\newblock Inception-v4, inception-resnet and the impact of residual connections
  on learning.
\newblock In {\em Proceedings of the AAAI conference on artificial
  intelligence}, volume~31, 2017.

\bibitem{szegedy2016rethinking}
Christian Szegedy, Vincent Vanhoucke, Sergey Ioffe, Jon Shlens, and Zbigniew
  Wojna.
\newblock Rethinking the inception architecture for computer vision.
\newblock In {\em Proceedings of the IEEE conference on computer vision and
  pattern recognition}, pages 2818--2826, 2016.

\bibitem{touvron2021training}
Hugo Touvron, Matthieu Cord, Matthijs Douze, Francisco Massa, Alexandre
  Sablayrolles, and Herv{\'e} J{\'e}gou.
\newblock Training data-efficient image transformers \& distillation through
  attention.
\newblock In {\em International conference on machine learning}, pages
  10347--10357. PMLR, 2021.

\bibitem{tran2022aisformer}
Minh Tran, Khoa Vo, Kashu Yamazaki, Arthur Fernandes, Michael Kidd, and Ngan
  Le.
\newblock Aisformer: Amodal instance segmentation with transformer.
\newblock {\em British Machine Vision Conference (BMVC)}, 2022.

\bibitem{vaswani2017attention}
Ashish Vaswani, Noam Shazeer, Niki Parmar, Jakob Uszkoreit, Llion Jones,
  Aidan~N Gomez, {\L}ukasz Kaiser, and Illia Polosukhin.
\newblock Attention is all you need.
\newblock {\em Advances in neural information processing systems}, 30, 2017.

\bibitem{vo2021aei}
Khoa Vo, Hyekang Joo, Kashu Yamazaki, Sang Truong, Kris Kitani, Minh-Triet
  Tran, and Ngan Le.
\newblock {{AEI}: Actors-Environment Interaction with Adaptive Attention for
  Temporal Action Proposals Generation}.
\newblock {\em BMVC}, 2021.

\bibitem{vo2022aoe}
Khoa Vo, Sang Truong, Kashu Yamazaki, Bhiksha Raj, Minh-Triet Tran, and Ngan
  Le.
\newblock Aoe-net: Entities interactions modeling with adaptive attention
  mechanism for temporal action proposals generation.
\newblock {\em International Journal of Computer Vision}, pages 1--22, 2022.

\bibitem{wang2022empirical}
Di~Wang, Jing Zhang, Bo~Du, Gui-Song Xia, and Dacheng Tao.
\newblock An empirical study of remote sensing pretraining.
\newblock {\em IEEE Transactions on Geoscience and Remote Sensing}, 2022.

\bibitem{wang2022advancing}
Di~Wang, Qiming Zhang, Yufei Xu, Jing Zhang, Bo~Du, Dacheng Tao, and Liangpei
  Zhang.
\newblock Advancing plain vision transformer towards remote sensing foundation
  model.
\newblock {\em IEEE Transactions on Geoscience and Remote Sensing}, 2022.

\bibitem{wang2020cse}
Fang Wang, Shihao Piao, and Jindong Xie.
\newblock Cse-hrnet: A context and semantic enhanced high-resolution network
  for semantic segmentation of aerial imagery.
\newblock {\em IEEE Access}, 8:182475--182489, 2020.

\bibitem{wang2021loveda}
Junjue Wang, Zhuo Zheng, Ailong Ma, Xiaoyan Lu, and Yanfei Zhong.
\newblock Loveda: A remote sensing land-cover dataset for domain adaptive
  semantic segmentation.
\newblock In J.~Vanschoren and S.~Yeung, editors, {\em Proceedings of the
  Neural Information Processing Systems Track on Datasets and Benchmarks},
  volume~1. Curran Associates, Inc., 2021.

\bibitem{wang2022novel}
Libo Wang, Rui Li, Chenxi Duan, Ce~Zhang, Xiaoliang Meng, and Shenghui Fang.
\newblock A novel transformer based semantic segmentation scheme for
  fine-resolution remote sensing images.
\newblock {\em IEEE Geoscience and Remote Sensing Letters}, 19:1--5, 2022.

\bibitem{wang2022unetformer}
Libo Wang, Rui Li, Ce~Zhang, Shenghui Fang, Chenxi Duan, Xiaoliang Meng, and
  Peter~M Atkinson.
\newblock Unetformer: A unet-like transformer for efficient semantic
  segmentation of remote sensing urban scene imagery.
\newblock {\em ISPRS Journal of Photogrammetry and Remote Sensing},
  190:196--214, 2022.

\bibitem{wang2021hierarchical}
Wenguan Wang, Tianfei Zhou, Siyuan Qi, Jianbing Shen, and Song-Chun Zhu.
\newblock Hierarchical human semantic parsing with comprehensive part-relation
  modeling.
\newblock {\em IEEE Transactions on Pattern Analysis and Machine Intelligence},
  44(7):3508--3522, 2021.

\bibitem{wang2018non}
Xiaolong Wang, Ross Girshick, Abhinav Gupta, and Kaiming He.
\newblock Non-local neural networks.
\newblock In {\em Proceedings of the IEEE conference on computer vision and
  pattern recognition}, pages 7794--7803, 2018.

\bibitem{Waqas2019}
Syed Waqas~Zamir, Aditya Arora, Akshita Gupta, Salman Khan, Guolei Sun, Fahad
  Shahbaz~Khan, Fan Zhu, Ling Shao, Gui-Song Xia, and Xiang Bai.
\newblock isaid: A large-scale dataset for instance segmentation in aerial
  images.
\newblock In {\em Proceedings of the IEEE/CVF Conference on Computer Vision and
  Pattern Recognition Workshops}, pages 28--37, 2019.

\bibitem{weiss2020remote}
Marie Weiss, Fr{\'e}d{\'e}ric Jacob, and Grgory Duveiller.
\newblock Remote sensing for agricultural applications: A meta-review.
\newblock {\em Remote sensing of environment}, 236:111402, 2020.

\bibitem{xia2023openearthmap}
Junshi Xia, Naoto Yokoya, Bruno Adriano, and Clifford Broni-Bediako.
\newblock Openearthmap: A benchmark dataset for global high-resolution land
  cover mapping.
\newblock In {\em Proceedings of the IEEE/CVF Winter Conference on Applications
  of Computer Vision}, pages 6254--6264, 2023.

\bibitem{xiao2018unified}
Tete Xiao, Yingcheng Liu, Bolei Zhou, Yuning Jiang, and Jian Sun.
\newblock Unified perceptual parsing for scene understanding.
\newblock In {\em Proceedings of the European Conference on computer vision
  (ECCV)}, pages 418--434, 2018.

\bibitem{xie2021segformer}
Enze Xie, Wenhai Wang, Zhiding Yu, Anima Anandkumar, Jose~M Alvarez, and Ping
  Luo.
\newblock Segformer: Simple and efficient design for semantic segmentation with
  transformers.
\newblock {\em Advances in Neural Information Processing Systems},
  34:12077--12090, 2021.

\bibitem{xu2020spatial}
Qingsong Xu, Xin Yuan, Chaojun Ouyang, and Yue Zeng.
\newblock Spatial--spectral ffpnet: Attention-based pyramid network for
  segmentation and classification of remote sensing images.
\newblock {\em arXiv preprint arXiv:2008.08775}, 2020.

\bibitem{xu2023rssformer}
Rongtao Xu, Changwei Wang, Jiguang Zhang, Shibiao Xu, Weiliang Meng, and
  Xiaopeng Zhang.
\newblock Rssformer: Foreground saliency enhancement for remote sensing
  land-cover segmentation.
\newblock {\em IEEE Transactions on Image Processing}, 32:1052--1064, 2023.

\bibitem{xue2022aanet}
Gunagkuo Xue, Yikun Liu, Yuwen Huang, Mingsong Li, and Gongping Yang.
\newblock Aanet: an attention-based alignment semantic segmentation network for
  high spatial resolution remote sensing images.
\newblock {\em International Journal of Remote Sensing}, 43(13):4836--4852,
  2022.

\bibitem{yamazaki2022vlcap}
Kashu Yamazaki, Sang Truong, Khoa Vo, Michael Kidd, Chase Rainwater, Khoa Luu,
  and Ngan Le.
\newblock Vlcap: Vision-language with contrastive learning for coherent video
  paragraph captioning.
\newblock In {\em 2022 IEEE International Conference on Image Processing
  (ICIP)}, pages 3656--3661. IEEE, 2022.

\bibitem{yamazaki2022vltint}
Kashu Yamazaki, Khoa Vo, Sang Truong, Bhiksha Raj, and Ngan Le.
\newblock Vltint: Visual-linguistic transformer-in-transformer for coherent
  video paragraph captioning.
\newblock {\em The Thirty-Seventh AAAI Conference on Artificial Intelligence},
  2023.

\bibitem{yang2018denseaspp}
Maoke Yang, Kun Yu, Chi Zhang, Zhiwei Li, and Kuiyuan Yang.
\newblock Denseaspp for semantic segmentation in street scenes.
\newblock In {\em Proceedings of the IEEE conference on computer vision and
  pattern recognition}, pages 3684--3692, 2018.

\bibitem{ye2019cross}
Linwei Ye, Mrigank Rochan, Zhi Liu, and Yang Wang.
\newblock Cross-modal self-attention network for referring image segmentation.
\newblock In {\em Proceedings of the IEEE/CVF conference on computer vision and
  pattern recognition}, pages 10502--10511, 2019.

\bibitem{you2019pixel}
Hongfeng You, Shengwei Tian, Long Yu, and Yalong Lv.
\newblock Pixel-level remote sensing image recognition based on bidirectional
  word vectors.
\newblock {\em IEEE Transactions on Geoscience and Remote Sensing},
  58(2):1281--1293, 2019.

\bibitem{yu2020context}
Changqian Yu, Jingbo Wang, Changxin Gao, Gang Yu, Chunhua Shen, and Nong Sang.
\newblock Context prior for scene segmentation.
\newblock In {\em Proceedings of the IEEE/CVF conference on computer vision and
  pattern recognition}, pages 12416--12425, 2020.

\bibitem{yuan2019segmentation}
Yuhui Yuan, Xiaokang Chen, Xilin Chen, and Jingdong Wang.
\newblock Segmentation transformer: Object-contextual representations for
  semantic segmentation.
\newblock {\em arXiv preprint arXiv:1909.11065}, 2019.

\bibitem{zhang2018context}
Hang Zhang, Kristin Dana, Jianping Shi, Zhongyue Zhang, Xiaogang Wang, Ambrish
  Tyagi, and Amit Agrawal.
\newblock Context encoding for semantic segmentation.
\newblock In {\em Proceedings of the IEEE conference on Computer Vision and
  Pattern Recognition}, pages 7151--7160, 2018.

\bibitem{zhang2021rest}
Qinglong Zhang and Yu-Bin Yang.
\newblock Rest: An efficient transformer for visual recognition.
\newblock {\em Advances in Neural Information Processing Systems},
  34:15475--15485, 2021.

\bibitem{zhao2017pyramid}
Hengshuang Zhao, Jianping Shi, Xiaojuan Qi, Xiaogang Wang, and Jiaya Jia.
\newblock Pyramid scene parsing network.
\newblock In {\em Proceedings of the IEEE conference on computer vision and
  pattern recognition}, pages 2881--2890, 2017.

\bibitem{zhao2018psanet}
Hengshuang Zhao, Yi~Zhang, Shu Liu, Jianping Shi, Chen~Change Loy, Dahua Lin,
  and Jiaya Jia.
\newblock Psanet: Point-wise spatial attention network for scene parsing.
\newblock In {\em Proceedings of the European conference on computer vision
  (ECCV)}, pages 267--283, 2018.

\bibitem{zhen2020joint}
Mingmin Zhen, Jinglu Wang, Lei Zhou, Shiwei Li, Tianwei Shen, Jiaxiang Shang,
  Tian Fang, and Long Quan.
\newblock Joint semantic segmentation and boundary detection using iterative
  pyramid contexts.
\newblock In {\em Proceedings of the IEEE/CVF Conference on Computer Vision and
  Pattern Recognition}, pages 13666--13675, 2020.

\bibitem{zheng2021rethinking}
Sixiao Zheng, Jiachen Lu, Hengshuang Zhao, Xiatian Zhu, Zekun Luo, Yabiao Wang,
  Yanwei Fu, Jianfeng Feng, Tao Xiang, Philip~HS Torr, et~al.
\newblock Rethinking semantic segmentation from a sequence-to-sequence
  perspective with transformers.
\newblock In {\em Proceedings of the IEEE/CVF conference on computer vision and
  pattern recognition}, pages 6881--6890, 2021.

\bibitem{zheng2020foreground}
Zhuo Zheng, Yanfei Zhong, Junjue Wang, and Ailong Ma.
\newblock Foreground-aware relation network for geospatial object segmentation
  in high spatial resolution remote sensing imagery.
\newblock In {\em Proceedings of the IEEE/CVF conference on computer vision and
  pattern recognition}, pages 4096--4105, 2020.

\bibitem{zhou2018unet++}
Zongwei Zhou, Md~Mahfuzur Rahman~Siddiquee, Nima Tajbakhsh, and Jianming Liang.
\newblock Unet++: A nested u-net architecture for medical image segmentation.
\newblock In {\em Deep Learning in Medical Image Analysis and Multimodal
  Learning for Clinical Decision Support: 4th International Workshop, DLMIA
  2018, and 8th International Workshop, ML-CDS 2018, Held in Conjunction with
  MICCAI 2018, Granada, Spain, September 20, 2018, Proceedings 4}, pages 3--11.
  Springer, 2018.

\bibitem{zhu2020deformable}
Xizhou Zhu, Weijie Su, Lewei Lu, Bin Li, Xiaogang Wang, and Jifeng Dai.
\newblock Deformable detr: Deformable transformers for end-to-end object
  detection.
\newblock {\em arXiv preprint arXiv:2010.04159}, 2020.

\end{thebibliography}

\end{document}